\documentclass{article}

\usepackage[preprint]{corl_2025} 

\usepackage{amsmath} 
\usepackage{amssymb}  
\usepackage{algorithm}
\usepackage{algpseudocode}
\usepackage{subfig}
\usepackage{graphicx}

\usepackage{multirow}
\usepackage{cuted}

\usepackage{booktabs}

\usepackage{multirow}
\usepackage{url}
\usepackage{makecell}
\usepackage{diagbox} 
\usepackage{bm}
\usepackage{tabularray}
\usepackage{color}

\usepackage[most]{tcolorbox}
\usepackage{titletoc}
\usepackage{setspace}
\usepackage{wrapfig}
\usepackage{xcolor}

\newcommand{\hide}[1]{}

\title{ MoDex: Planning High-Dimensional Dexterous Control via Learning Neural Internal Models }

\author{
Tong Wu\textsuperscript{$1*$},\quad
Shoujie Li\textsuperscript{$1*$},\quad
Chuqiao Lyu\textsuperscript{$1$},\quad
Kit-Wa Sou\textsuperscript{$1$},\quad \\
\vspace{0.2cm}
\textbf{
Wang-Sing Chan\textsuperscript{$1$},\quad 
Wenbo Ding\textsuperscript{$1$$\dag$}
} \\
\vspace{0.1cm}
\textsuperscript{$*$} Equal Contributions, 
\textsuperscript{$\dag$} Corresponding Author \\
\textsuperscript{$1$} Shenzhen International Graduate School, Tsinghua University \\
}

\begin{document} 

\maketitle


\begin{figure}[h]
  \centering
  \captionsetup{font=small}
  \includegraphics[width=0.95\linewidth]{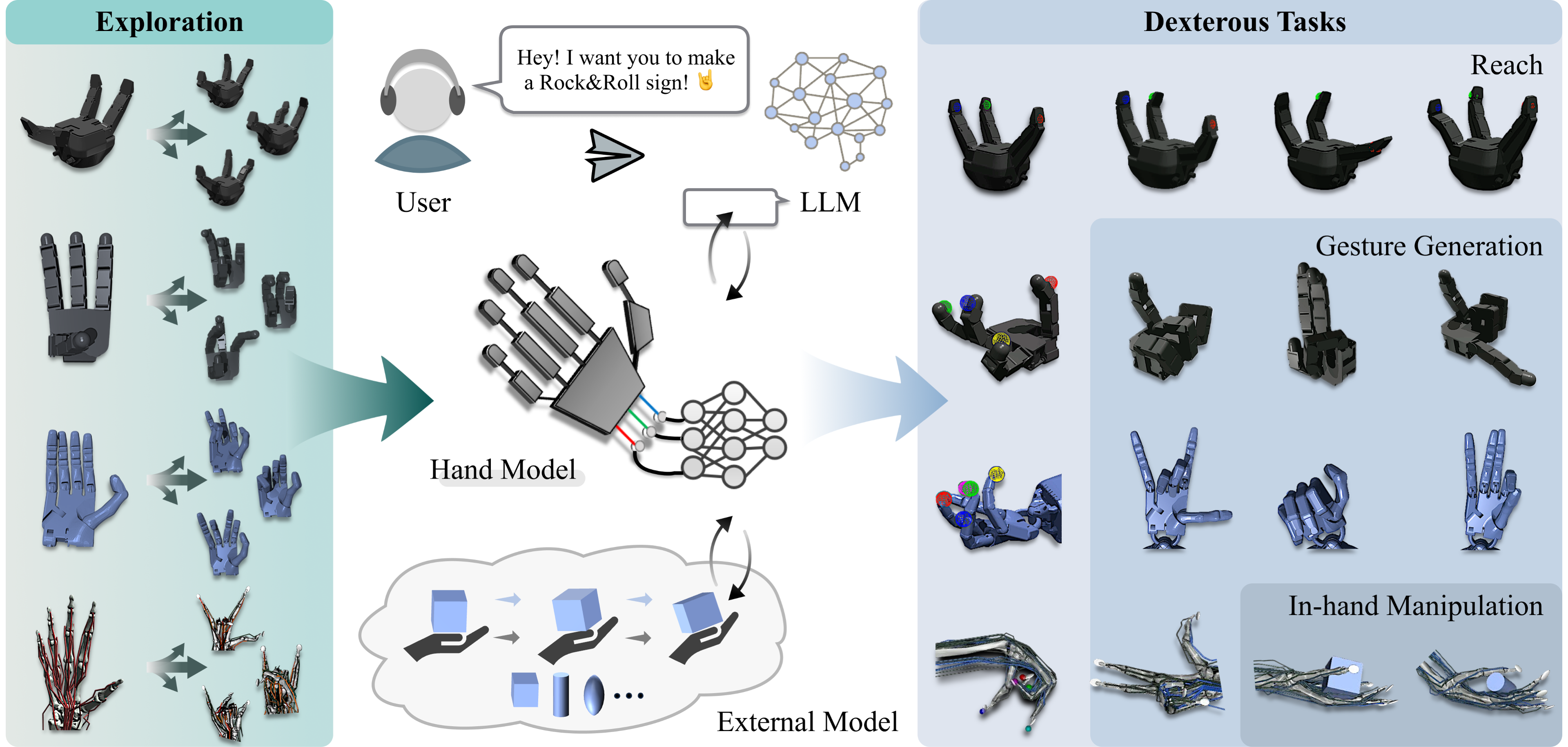}
  \caption{\textbf{\textbf{MoDex}}. Our framework learns neural internal models to represent various dexterous hands. MoDex enables precise control in high-dimensional action space with an internal model, the generation of diverse gestures by integrating the hand model with the LLM, and data-efficient in-hand manipulation via learning factorized dynamics model.}
  \label{fig:teaser}
  \vskip -0.2in
\end{figure}

\begin{abstract}
Controlling hands in high-dimensional action space has been a longstanding challenge, yet humans naturally perform dexterous tasks with ease. In this paper, we draw inspiration from the concept of \textit{internal model} exhibited in human behavior and reconsider dexterous hands as learnable systems. Specifically, we introduce MoDex, a framework that includes a couple of neural networks (NNs) capturing the dynamical characteristics of hands and a bidirectional planning approach, which demonstrates both training and planning efficiency. To show the versatility of MoDex, we further integrate it with an external model to manipulate in-hand objects and a large language model (LLM) to generate various gestures in both simulation and real world. Extensive experiments on different dexterous hands address the data efficiency in learning a new task and the transferability between different tasks. 
\end{abstract}


\section{Introduction}
\label{sec:intro}

The broad functional potential of dexterous hands has motivated extensive research in areas such as dexterous grasping~\cite{wang2023dexgraspnet, xu2023unidexgrasp,mandikal2021learning}, in-hand manipulation~\cite{handa2023dextreme, doi:10.1126/scirobotics.adc9244, touch-dexterity, qi2023general} and even musculoskeletal manipulation~\cite{wang2022myosim, caggiano2022myosuite}. 
Despite this progress, controlling high degree-of-freedom (DoF) hands remains a major challenge, as the data required to train a policy grows with the action dimension~\cite{caggiano23myochallenge}. 
Deep reinforcement learning (DRL), a trial-and-error-based approach, struggles with this problem due to its need for an unreasonably large number of interactions in high-dimensional settings~\cite{caggiano2023myodex}.

One common approach to addressing this challenge is reducing the dimensionality of the action space by leveraging \textit{synergy}, a concept from neuroscience that describes the coordinated control of multiple actuators~\cite{grillner1985neurobiological}. 
Previous works have applied this idea to enhance DRL efficiency by exploiting the action redundancy in specific tasks through Manifold learning~\cite{zuo2024self} and covariance analysis~\cite{berg2023sar, he2024dynsyn}. 
While synergy-based methods can accelerate learning, they inherently shape a task-related reduced action space, leaving the transferring of muscle synergies largely unexplored~\cite{berg2023sar}. 
Additionally, synergies themselves vary significantly depending on the task; for instance, chopstick using and cube rotating differ in both the pattern and quantity of synergies (See Appendix~\ref{sup:syn} for more details). These characteristics may affect the performance across different tasks.


The \textit{internal model}, another concept from neuroscience, suggests that rather than mapping observations directly to actuator controls, the nervous system relies on a pair of models: a forward model and an inverse model~\cite{kawato1999internal, tin2005internal}. 
The forward model predicts the outcomes of actions based on observed dynamics, while the inverse model infers the necessary actuator commands to achieve a target state. 
As an example, humans demonstrate remarkable efficiency in dexterous control over hundreds of muscles~\cite{egger2019internal, angelaki2004neurons}. 
This raises the question: can we endow robots with similar high-dimensional control capabilities using the internal model? 

In this paper, we investigate whether the internal model can facilitate high-dimensional dexterous control. 
Unlike traditional model-based approaches that focus on learning environment dynamics~\cite{nagabandi2018neural, shi2024robocraft}, our approach, \textbf{MoDex}, addresses that the dexterous hand itself should be studied as an independent system. 
Specifically, a pair of NN-based forward model and inverse model are trained through random exploration, which provides robots with an understanding of hand dynamics. 
To accelerate the decision-making process, we propose a bidirectional planning method by integrating the neural internal model with a Cross-Entropy Method~\cite{rubinstein2004cross} (CEM) planner. 
Draw inspiration from neuroscience experiments, we evaluate MoDex on fingertip control to reach target positions using four dexterous hands in simulation (as shown in Appendix~\ref{sup:hands}). 
Results show that our approach is significantly more data efficient compared to model-free methods while also demonstrating faster planning compared to model-based baselines.

To further demonstrate its versatility, we utilize the pretrained neural internal model as a plug-and-play module for in-hand manipulation and few-shot gesture generation in both simulation and real world. 
In the first case, we factorize the environment dynamics into the pretrained internal model and an external dynamics model, leading to improved data efficiency in object reorientation. 
In the second case, we integrate MoDex to a large language model (LLM) by prompting it to generate cost functions. 
Our approach successfully generates a variety of gestures, such as ``Scissorshand" and ``Rock\&Roll".
These examples indicate MoDex's potential for diverse high-dimensional control tasks.\footnote{\href{https://mo-dex.github.io/}{Project website: https://mo-dex.github.io/}}


    



\section{Related Work}
\label{sec:rw}

\subsection{Internal Model}
The concept of internal model was first introduced to explicitly represent control systems~\cite{francis1976internal}. In biological motor control, growing evidence suggests that the human body maintains an internal model to predict and adapt movements~\cite{egger2019internal, angelaki2004neurons}. Since real-time sensory feedback is often unreliable due to occlusions or noise, motor planning primarily relies on an internal model constructed from prior perception~\cite{kawato1999internal}. For instance, when vision is obstructed or the environment is dark, humans still accurately estimate hand position, a phenomenon best explained by an internal model~\cite{wolpert1995internal}. Similarly, motor adaptation experiments reveal that humans adjust to external disturbances by continuously updating their internal models and, even after the disturbances are removed, momentarily retain compensatory responses~\cite{shadmehr1994adaptive}. 
Motivated by these insights, we propose learning neural network-based internal models to enable effective planning in high-dimensional dexterous control, emulating the predictive and adaptive mechanisms observed in human motor learning.

\subsection{Dexterous Manipulation} 
Dexterous manipulation has garnered significant attention due to the remarkable performance of model-free RL. However, effectively controlling a dexterous hand remains challenging, particularly for in-hand manipulation, which requires precise reorientation of objects. To address this, prior works have leveraged vision~\cite{doi:10.1126/scirobotics.adc9244}, touch~\cite{touch-dexterity}, or a combination of both~\cite{yuan2023robot}. Musculoskeletal hands further expand the scope of dexterous manipulation by introducing higher DoFs and more human-like control capabilities~\cite{caggiano2022myosuite, caggiano2023myodex}. However, the increased action dimensionality poses a challenge for efficient learning. Synergy-based approaches mitigate this issue by exploiting task~\cite{berg2023sar} and embodiment redundancies~\cite{zuo2024self, he2024dynsyn}, reducing the effective control complexity. Beyond model-free RL, deep model-based methods learn system dynamics using neural networks~\cite{nagabandi2018neural}. When integrated with powerful planning techniques, these methods offer both efficiency and flexibility~\cite{nagabandi2020deep}. While our approach falls under model-based learning, it differs by modeling the hand as an independent system, introducing a hierarchical perspective to dexterous control.
\section{Method}
\label{sec:method}

\subsection{Learning Neural Internal Models}
Controlling a hand is exceptionally complex, yet humans manage it effortlessly, suggesting the presence of internal model planning~\cite{ito2008control}. 
This observation underscores the importance of explicitly modeling the body system for dexterous control. 
Accordingly, we propose to learn a neural internal model to enable high-dimensional dexterous control. 
The neural internal model consists of:
\subsubsection{Forward Model for Dexterous Hand Dynamics} 
To achieve precise high-dimensional control, we introduce a neural forward model that approximates the dynamics of the dexterous hand. Specifically, given the current state $\boldsymbol{s}_t$ and action $\boldsymbol{a}_t$, the forward model predicts the next state:
\begin{equation} \label{eq:1} \hat{\boldsymbol{s}}_{t+1} = f_{\theta}(\boldsymbol{s}_t, \boldsymbol{a}_t), \end{equation}
where $\boldsymbol{s}_t, \hat{\boldsymbol{s}}_{t+1} \in \mathbb{R}^{H}$ represents the hand state and the imagined next state respectively, which we define as the positions of all fingertips. $\boldsymbol{a}_t \in \mathbb{R}^{K}$ denotes the action generated by actuators, with $K$ being the action dimension. $f_{\theta}(\cdot)$ is a parametric function modeled as a multi-layer perceptron (MLP) with parameters $\theta$.

To train the forward model, we collect a dataset of transitions $\mathcal{D} = \{(\boldsymbol{s}_t, \boldsymbol{a}_t, \boldsymbol{s}_{t+1})_i\}$ using random exploration. 
The forward model is then trained to minimize a multi-step prediction loss:
\begin{equation} L_\text{forward} = \sum_{i=0}^{S} \alpha^i ||\boldsymbol{s}_{t+i+1} - f_{\theta}(\hat{\boldsymbol{s}}_{t+i}, \boldsymbol{a}_{t+i})||^2, \end{equation}
where $S$ is the prediction horizon, controlling how many future steps are predicted, and $\alpha$ is a discount factor, weighting short-horizon errors more than long-horizon errors. $\hat{\boldsymbol{s}}_t$ is set to $\boldsymbol{s}_t$ for initial alignment. This multi-step objective prevents compounding errors and encourages the model to generalize beyond short-term predictions, making it more reliable for downstream planning tasks.



\begin{figure*}[]
\centering
\captionsetup{font=small}
\includegraphics[width=0.95\textwidth]{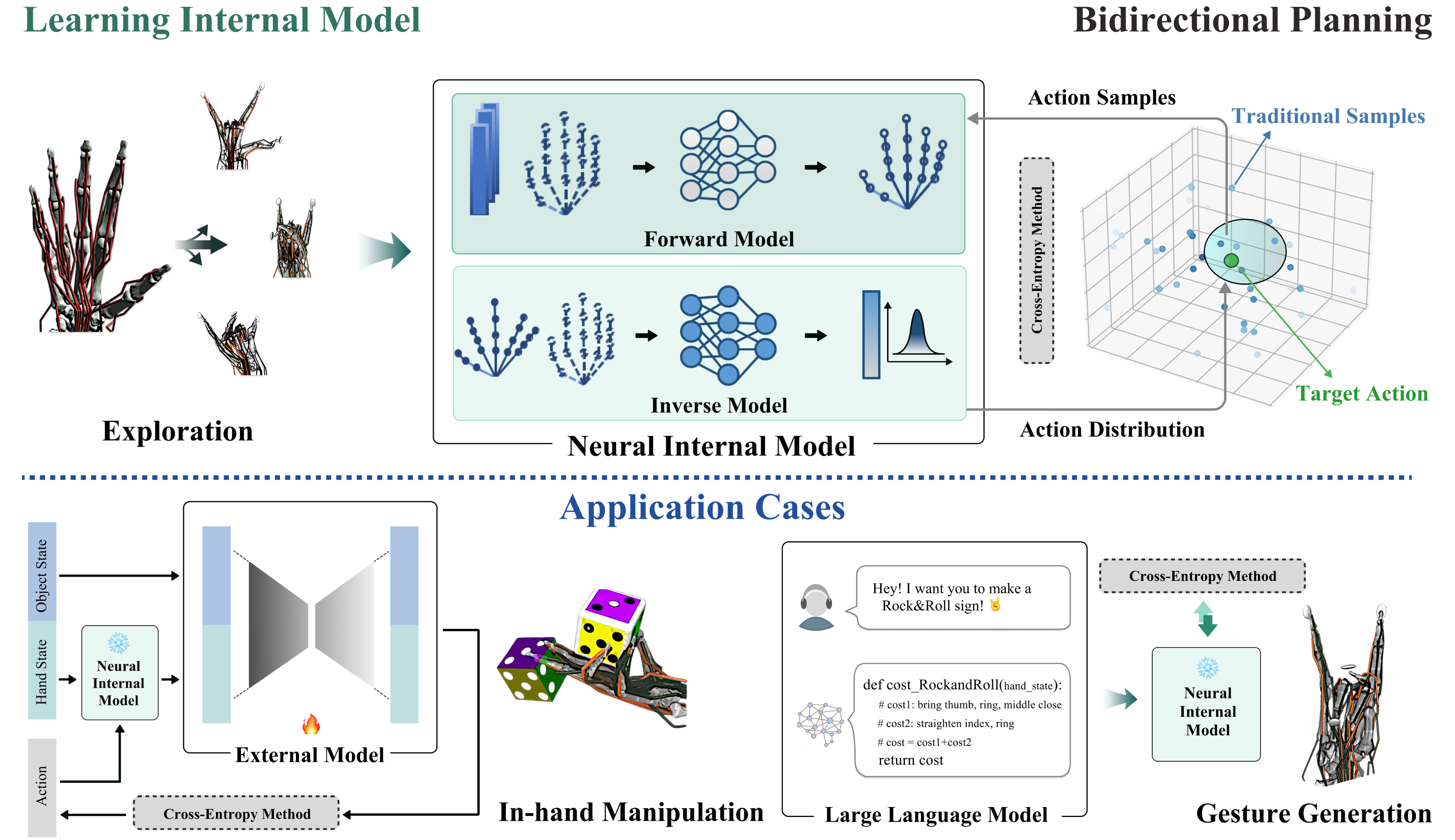}
\caption{\textbf{Method overview.} We first explore the action space and collect dynamics data to train the internal model. Using these models, we employ CEM-based bidirectional planning to optimize actions. \textbf{Applications.} We decompose the system dynamics into a hand model and an external model, enhancing learning efficiency in in-hand manipulation task. Additionally, we leverage LLM to generate a cost function from textual inputs, guiding action optimization for gesture generation.} 
\label{fig:method}
\vskip -0.2in
\end{figure*}


\subsubsection{Inverse Model for Action Generation}

The inverse model operates in the reverse direction of the forward model, predicting an action that moves the hand toward a desired target state. Formally, this process is defined as:
\begin{equation}
\boldsymbol{\hat{a}_t} = g_{\phi}(\boldsymbol{s}_t,\boldsymbol{s}_T),
\end{equation}
where $g_{\phi}$ is an MLP parameterized by $\phi$, and $\boldsymbol{s}_T \in \mathbb{R}^{H}$ represents the target hand state. 
However, action prediction is inherently ambiguous since multiple valid action sequences may lead to the same target state. Additionally, the accuracy of the predicted action depends on the information contained in the state. To address this, we model the predicted action as a Gaussian distribution rather than a deterministic output. Specifically, we define:
\begin{equation}
    \boldsymbol{\hat{a}_t} \sim \mathcal{N} \big( g_{\phi}(\boldsymbol{s}_t,\boldsymbol{s}_T), \Sigma) \big),
\end{equation}
where $\Sigma$ is a diagonal covariance matrix, and $\boldsymbol{\sigma}$, its square root of diagonal vector is estimated from the training dataset as:
\begin{equation}
    \boldsymbol{\sigma} = \mathbb{E}_{(\boldsymbol{s}_t, \boldsymbol{a}_t, \boldsymbol{s}_T) \sim \mathcal{D}} \Big[ | \boldsymbol{a}_t - g_{\phi}(\boldsymbol{s}_t,\boldsymbol{s}_T) | \Big].
\end{equation}
This distribution is then utilized to generate a set of action samples, which are refined through additional optimization as described in the following section.

We train the inverse model on the collected dataset $\mathcal{D}$ and use L1 loss function:
\begin{equation}
    L_\text{inverse} = \mathbb{E}_{(\boldsymbol{s}_t, \boldsymbol{a}_t, \boldsymbol{s}_T) \sim \mathcal{D}} \Big[ |g_{\phi}(\boldsymbol{s}_t,\boldsymbol{s}_{t+t_0}) - \boldsymbol{a}_t| \Big],
\end{equation}
where $t_0$ is a hyperparameter controlling the target timestamp shift, which we set to 1 in this work.


\subsection{Bidirectional Planning Strategy}
When executing complex movements, the central nervous system (CNS) decomposes motion into an initial rapid movement followed by corrective sub-movements~\cite{woodworth1899accuracy,elliott2001century}. Inspired by this, our bidirectional planning strategy first generates an initial action distribution using the inverse model, then refines it iteratively.

Given a target hand state, the inverse model produces an action distribution, which is extended over a planning horizon $T_0$ to initialize the sampling process. A CEM planner iteratively refines the action sequence by sampling candidate actions from the distribution, predicting their outcomes via the forward model, selecting the top-performing samples, and updating the distribution based on their mean and variance~\cite{rubinstein2004cross}. For sequential tasks, we further integrate Model Predictive Control (MPC)~\cite{nagabandi2018neural}, which updates plans based on real-time observations. See Appendix~\ref{sup:biplan} for more details.

\section{Applications of Pretrained Internal Model}
To demonstrate the versatility of a pretrained neural internal model, we explore two application cases: factorized dynamics learning for in-hand manipulation and few-shot gesture generation.
\subsection{Factorized Dynamics Learning}
Instead of modeling the hand and external environment jointly for in-hand manipulation, we adopt a hierarchical factorization. The complete system dynamics are decomposed into (i) a pretrained internal model and (ii) an external dynamics model capturing interactions with objects.

Given an action, we first predict the next hand state using the pretrained model (Equation~\ref{eq:1}), which serves as an implicit prior since humans acquire such models through experience. The external dynamics model then updates the prediction by incorporating the interaction with the object:
\begin{equation} \label{eq:4} \boldsymbol{s}_{t+1},\boldsymbol{x}_{t+1} = f_{\psi}(\hat{\boldsymbol{s}}_{t+1}, \boldsymbol{x}_t), \end{equation}
where $\boldsymbol{x}_t$ and $\boldsymbol{x}_{t+1}$ are the external states before and after interaction, while $\hat{\boldsymbol{s}}_{t+1}$ and $\boldsymbol{s}_{t+1}$ denote the hand states in imagination and after interaction. The external model $f_{\psi}$, parameterized by an MLP, learns the dynamics of interaction. This factorization can directly eliminate the action dimension from the dynamics model (See Appendix~\ref{sup:facdyn} for analysis).
To learn a specific task, we employ an online adaptive learning strategy~\cite{nagabandi2020deep} that iteratively rolls out actions, collects new transition data, and updates the models. Multi-step loss is applied to enhance long-horizon prediction robustness.

\subsection{Few-shot Gesture Generation}
We formulate gesture generation from text input as an optimization problem, leveraging LLMs' common knowledge and coding ability to provide the objective function without post-training.

Given a linguistic gesture request, we require an LLM to generate a heuristic function that quantifies the cost of the current hand state to the target gesture. 
For example, the ``OK" gesture with a five-fingered hand can be described by:
\begin{equation} \mathcal{J}(\boldsymbol{s}) = - \Vert \boldsymbol{s}_{0} - \boldsymbol{s}_{1} \Vert_2 + \sum_{i>2}(\boldsymbol{s}_i \cdot \hat{\boldsymbol{n}}_i), \end{equation}
where $\boldsymbol{s}_i$ represents the $i$-th fingertip position, and $\hat{\boldsymbol{n}}_i$ denotes the preferred direction for straightening each finger. It encourages thumb-index contact while keeping other fingers extended. 
This cost function is presented in python function as shown in Figure~\ref{fig:method}. 
To enable few-shot ability, we integrate in-context learning~\cite{huang2023voxposer}, supplying the LLM with multiple pre-defined cost functions as exemplars. 
Finally, MoDex is used to generate the action for the target gesture, where the LLM-generated cost function serves as the objective.

\section{Experiments}
\label{sec:experiment}

\subsection{Fingertip Reach Experiment}
\label{exp:reach}

\begin{table}[]
\centering
\captionsetup{font=small}
\caption{\textbf{Evaluation on fingertip reach task.}
We exclude model-free baselines from this quasi-static setting due to their inability to succeed. Experimental results demonstrate that our method achieves the lowest fingertip reach error while requiring the fewest planning samples, highlighting both its accuracy and sample efficiency.
}
\resizebox{\linewidth}{!}{%
\begin{tabular}{l|ccc|ccc|ccc|ccc}
\toprule
\multirow{2}{*}{\textbf{Methods}} & \multicolumn{3}{c|}{Robotiq} & \multicolumn{3}{c|}{Allegro} & \multicolumn{3}{c|}{Shadowhand} & \multicolumn{3}{c}{Myohand} \\  
\cmidrule(lr){2-13} 
& \textbf{S.R. \textuparrow} & \textbf{R.E. \textdownarrow} & \textbf{P.S. \textdownarrow} & \textbf{S.R. \textuparrow} & \textbf{R.E. \textdownarrow} & \textbf{P.S. \textdownarrow} & \textbf{S.R. \textuparrow} & \textbf{R.E. \textdownarrow} & \textbf{P.S. \textdownarrow} & \textbf{S.R. \textuparrow} & \textbf{R.E. \textdownarrow} & \textbf{P.S. \textdownarrow} \\ 
\midrule
\multicolumn{13}{c}{\textbf{Quasi-static Setting}} \\ 
\midrule
FM+RS   &    79     &  0.80   &  10k   &     49     &   1.47   &   10k   &     22      &  1.78    &   10k   & 34    & 1.58 &  10k \\
FM+BGD   &    82     &  0.76    &   38.4k   &    66     &  1.33    &   38.4k   &      45     &   1.50   &   38.4k   & 66    & 1.13 &  38.4k \\
FM+CEM  &     88    &   0.77   &   1.2k   &     70   &   1.34   &   1.2k   &       \textbf{\textbf{55}}     &   \textbf{\textbf{1.45}}   &   1.2k   & 77    &  1.01  & 2.0k \\
Ours    &    \textbf{\textbf{91}}     &   \textbf{\textbf{0.72}}   &   \textbf{\textbf{0.8k}}   &     \textbf{\textbf{72}}    &  \textbf{\textbf{1.32}}    &   \textbf{\textbf{1.0k}}   &    54     &   1.47   &   \textbf{\textbf{1.0k}}   &  \textbf{\textbf{87}}     &  \textbf{\textbf{0.87}}   &  \textbf{\textbf{2.0k}} \\ 
\midrule
\multicolumn{13}{c}{\textbf{Sequential Setting}} \\ 
\midrule
SAC      &    89     &   0.74   &   \textendash   &     50    &   1.52   &   \textendash   &      38      &   1.65   &   \textendash   &    62    &  1.52      &       \textendash                \\
SAR      &    8     &   2.03   &   \textendash   &     0    &   2.44   &   \textendash   &      2      &   2.38   &   \textendash   &    0    &  2.54      &       \textendash                \\
FM+RS   &    84     &  0.76  &  50k  &     53     &   1.50  &   50k   &    45    &  1.60  &  50k    &  7   & 2.50 & 50k  \\
FM+BGD   &    80     &   0.83   &  38.4k    &    62     &  1.38    &  38.4k     &     43    &   1.56   &   38.4k    &  85  & 1.20 & 38.4k  \\
FM+CEM  &  90    &   0.74   &  2.0k  &   86    &  1.19  & 2.0k   &   50     &  1.52  &  2.4k   & 77   &  1.25   &  2.4k\\
Ours    &    \textbf{\textbf{96}}    &   \textbf{\textbf{0.51}}   &  \textbf{\textbf{1.2k}}   &  \textbf{\textbf{89}}   &  \textbf{\textbf{1.16}}    &   \textbf{\textbf{1.2k}}   &   \textbf{\textbf{56}}   &   \textbf{\textbf{1.48}}  &  \textbf{\textbf{1.2k}}   &    \textbf{\textbf{88}}   &  \textbf{\textbf{1.18}}   & \textbf{\textbf{1.8k}} \\                    
\bottomrule
\end{tabular}
}
\label{tab:combined}
\vskip -0.1in
\end{table}

\begin{figure*}[]
\centering
\captionsetup{font=small}
\includegraphics[width=0.95\textwidth]{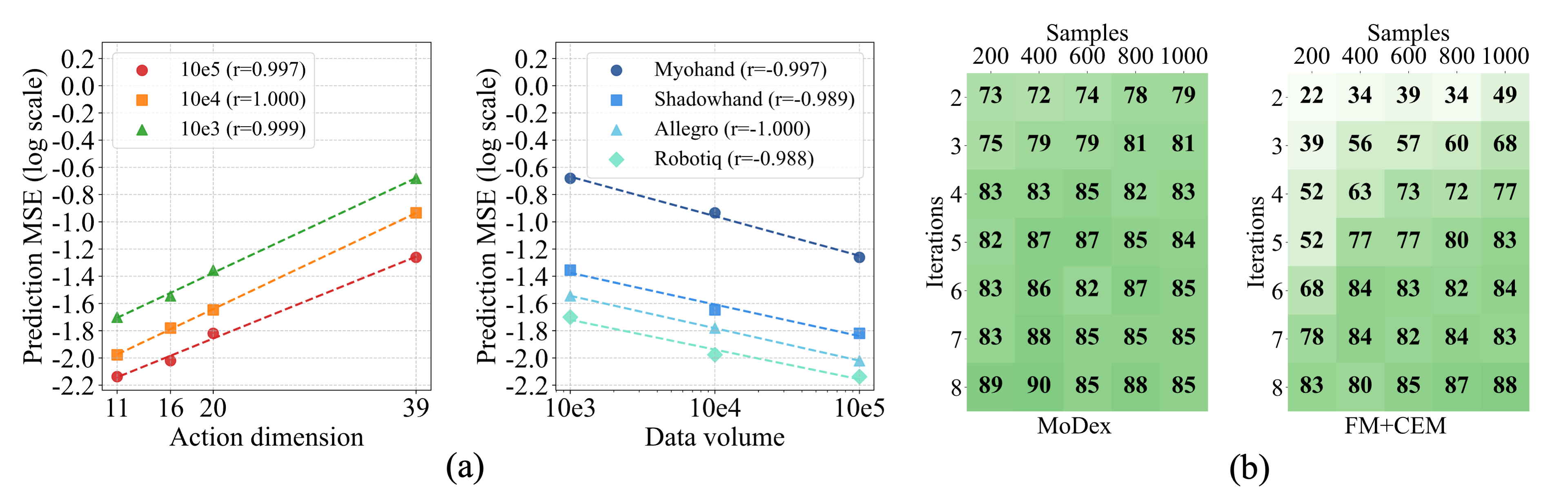}
\caption{\textbf{Ablation studies.} \textbf{(a)} Ablation on forward model. We find that to maintain the same prediction error the training data grows exponentially with the action dimension. \textbf{(b)} Ablation on planning method. Results demonstrate the inverse process of MoDex can significantly reduce planning samples and iterations.} 
\label{fig:ablation}
\vskip -0.2in
\end{figure*}

\textbf{Task and evaluation settings.} 
We evaluate a fingertip reach task in simulation, inspired by neuroscience studies, using four dexterous hands from IsaacGym~\cite{makoviychuk2021isaac} and Myosuite~\cite{caggiano2022myosuite}. The goal is to move fingertips to target positions (see Appendix~\ref{sup:reachresults} for more details). We consider two settings: \textbf{(1)} \textit{Quasi-static} (single-step), and \textbf{(2)} \textit{Sequential} (multi-step); see Appendix~\ref{sup:setting}. Performance is measured over 100 episodes using: (1) \textbf{S.R.} (Success Rate) – percentage of episodes meeting a distance threshold, (2) \textbf{R.E.} (Reach Error) – average fingertip-target distance, and (3) \textbf{P.S.} (Planning Samples) – samples used per planning step.

\textbf{Baselines.} 
We compare MoDex against two categories of baselines: 
(1) Model-free methods – Soft Actor-Critic~\cite{haarnoja2018soft} (\textbf{SAC}) and the synergy-based approach~\cite{berg2023sar} (\textbf{SAR}); (2) Model-based methods – Random Shooting (\textbf{FM+RS}), Batch Gradient Descent (\textbf{FM+BGD}), and the Cross-Entropy Method~\cite{nagabandi2020deep} (\textbf{FM+CEM}). 

\textbf{Result analysis.} As shown in Table~\ref{tab:combined}, MoDex outperforms both model-free and most model-based baselines while demonstrating superior planning efficiency. Moreover, model-based methods require only 1\% and 10\% of the training data used by model-free methods in the quasi-static and sequential settings, respectively, highlighting their learning efficiency (Appendix~\ref{sup:reachresults}). Notably, SAR struggles in this task, suggesting that synergy-based approaches may be less effective in low-redundancy tasks. We provide a synergy analysis in Appendix~\ref{sup:syn} for further investigation.

\subsection{Ablation Studies}
\textbf{Relation of action dimension and data volume.} 
We analyze the effects of action dimension and training data volume on forward model performance using MSE as the prediction error with a 10:1 train/evaluation ratio.
As shown in Figure~\ref{fig:ablation}(a), log prediction error decreases linearly with data volume but increases linearly with action dimension. This suggests that maintaining consistent performance requires data volume to scale exponentially with action dimension.

\textbf{Why MoDex outperforms other model-based methods?} A key question is how bidirectional planning—the primary distinction between MoDex and other model-based methods—contributes to its higher success rate, especially in comparison to CEM. 
As shown in Figure~\ref{fig:ablation}(b), FM+CEM achieves performance comparable to MoDex when provided with sufficiently large sample sizes and iterations. However, under tighter planning budgets, MoDex exhibits a more significant advantage. This suggests that the inverse model offers a strong initialization for the subsequent forward-planning process (Figure~\ref{fig:method}), while the forward model ultimately determines the asymptotic performance. More results can be found in Appendix~\ref{sup:reachresults}.

\subsection{In-Hand Manipulation}

\begin{figure}[]
\centering
\captionsetup{font=small}
\includegraphics[width=1\textwidth]{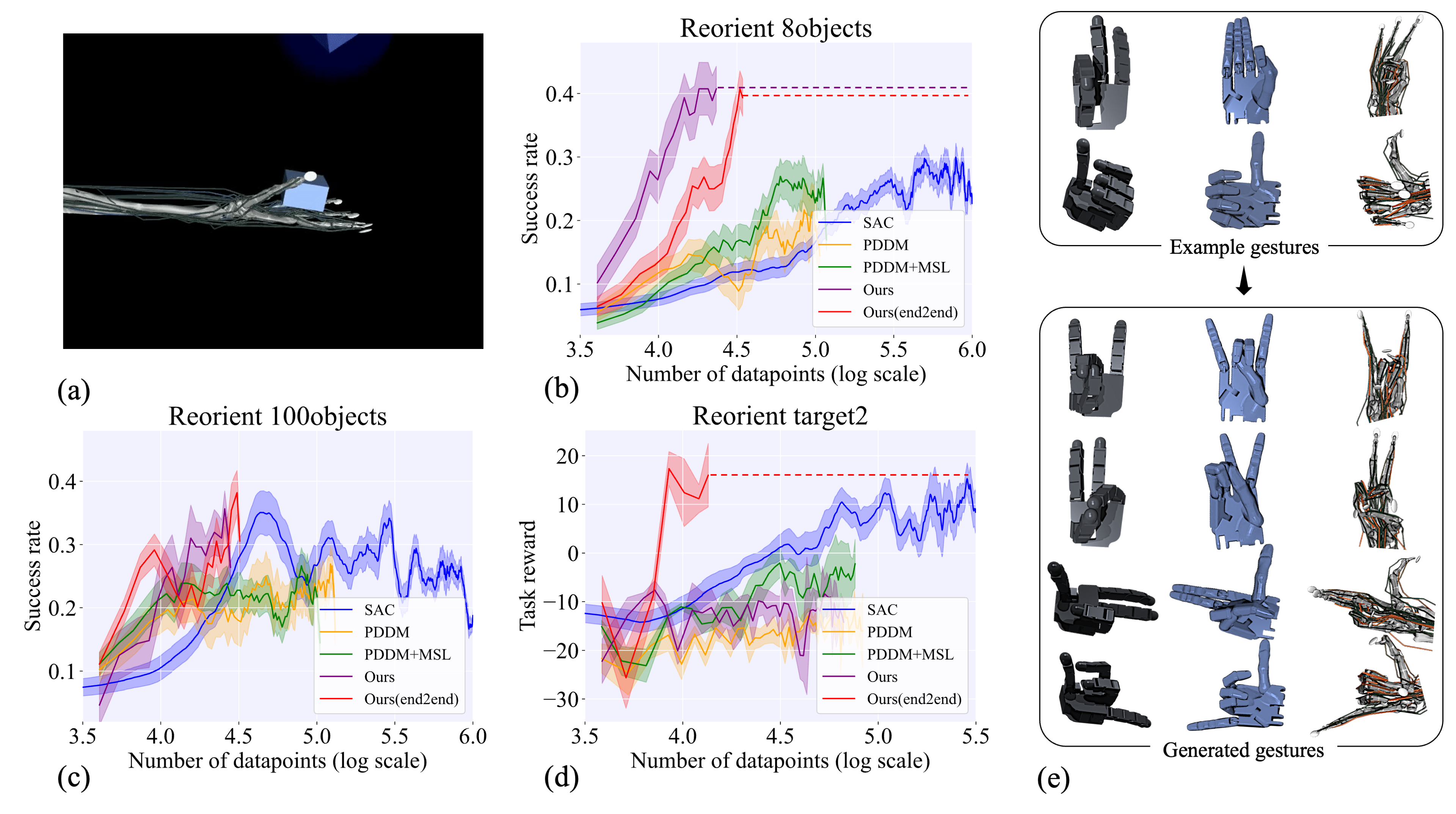}
\caption{\textbf{Application cases.} \textbf{(a)} In-hand manipulation with MyoHand. \textbf{(b–d)} Quantitative evaluation on three in-hand manipulation tasks shows that MoDex achieves the highest success rates with the fewest data. \textbf{(e)} Gesture generation results: given two example gestures, MoDex generates four novel gestures using the LLM.} 
\label{fig:applications}
\vskip -0.2in
\end{figure}


\textbf{Task and evaluation settings.}
We evaluate factorized dynamics learning to in-hand manipulation tasks using Myohand~\cite{caggiano2022myosuite}: \textbf{Reorient target2}, \textbf{Reorient 8objects}, \textbf{Reorient 100objects} (Figure~\ref{fig:applications}(a)). To demonstrate the plug-and-play ability, we utilized the internal model pretrained in sequential setting of fingertip reach experiment.

\textbf{Baselines.}
We compare our method against three baselines: 1) Soft Actor-Critic~\cite{haarnoja2018soft} (\textbf{SAC}), 2) \textbf{PDDM}~\cite{nagabandi2020deep}, which learns a single dynamics model for the entire manipulation system, and 3) \textbf{PDDM+MSL}, an extension of PDDM incorporating a multi-step loss to improve long-term prediction accuracy. We also evaluate the end-to-end training version of our method \textbf{Ours (end2end)} by unfreezing the pretrained model.

\textbf{Result analysis.}
As shown in Figure~\ref{fig:applications}(b,c), our method outperforms all baselines on Reorient 8objects and 100objects with significantly less data than model-free RL. While other model-based methods are more data-efficient than RL, they underperform our approach, which focuses solely on training the external model. Our method also matches or exceeds its end-to-end variant on Reorient 8objects. However, it fails on Reorient target2 due to potential error accumulation from model composition, which can be alleviated by unfreezing the pretrained internal model (Figure~\ref{fig:applications}(d)).

\subsection{Gesture Generation Study} 


\textbf{Task setting.}
We evaluate MoDex’s capability in gesture generation using AllegroHand, ShadowHand, and MyoHand, excluding Robotiq due to its unsuitability for gesture posing. To assess transferability, we directly employ the pretrained internal model from the quasi-static setting of the fingertip reach experiment. Additionally, to demonstrate few-shot learning, we utilize GPT-4 as the core model for generating cost functions, providing only two example cost functions in the system prompt (see the top of Figure~\ref{fig:applications}(e)).

\textbf{Result analysis.} As shown in Figure~\ref{fig:applications}(e), the generated gestures include diverse forms such as Rock\&Roll, Scissorhands, Finger Gun, and Calling. We provide the example cost functions and generated cost functions of Allegrohand in Appendix~\ref{sup:gesture}. The successful generation of gestures and in-hand manipulation using a pretrained internal model highlights MoDex's transferability. 

\subsection{Real-World Deployment}
\label{sup: }

\begin{figure}[]
\captionsetup{font=small}
\centering
\includegraphics[width=\textwidth]{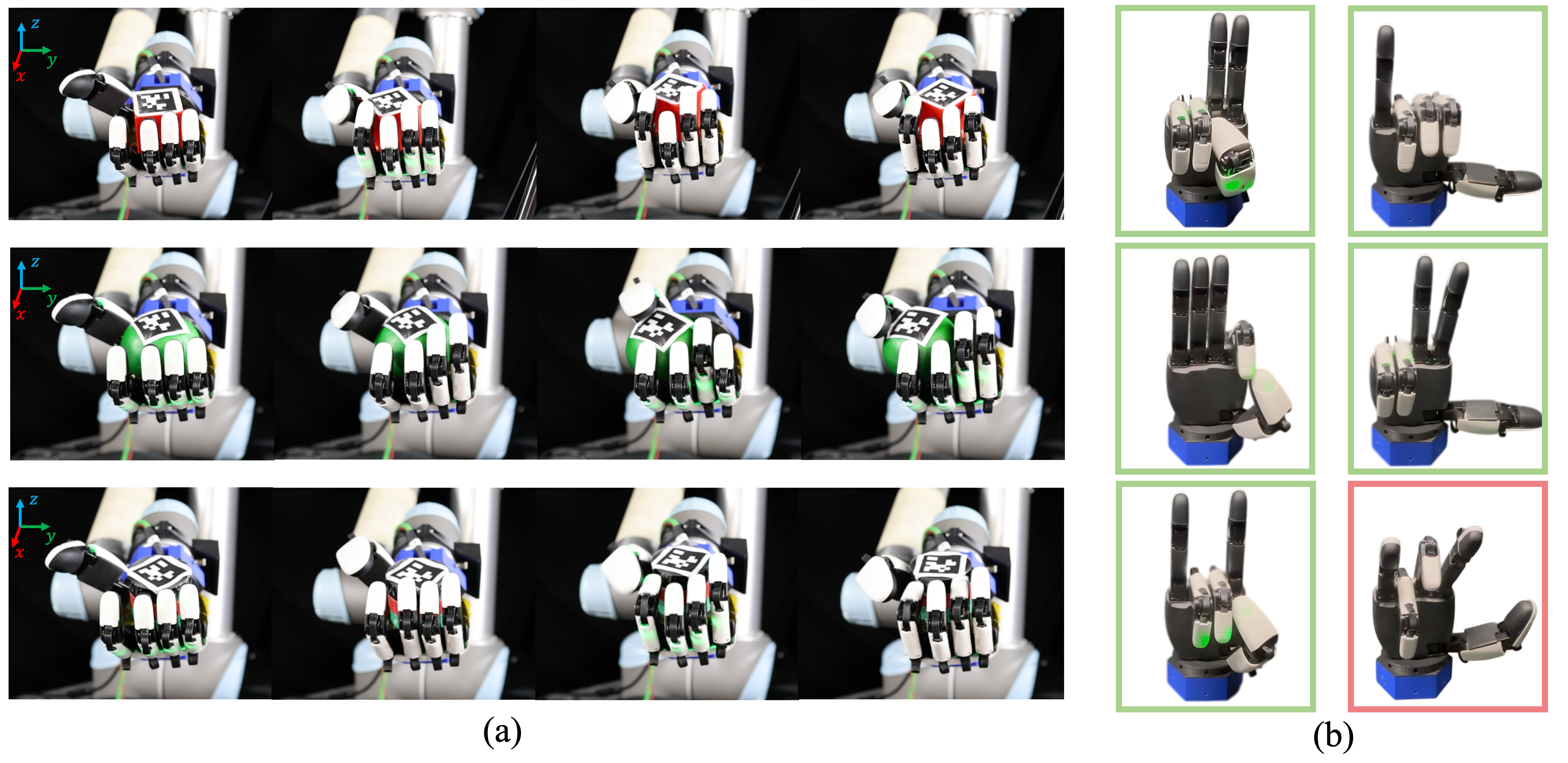}
\caption{\textbf{Real-World deployment.} \textbf{(a)} We deploy MoDex to perform in-hand rotation of three objects (cube, orange, and cylinder) around the $z$-axis. \textbf{(b)} MoDex generates diverse hand gestures in the real world. \textcolor{green!50!black}{Green} indicates successful executions, while \textcolor{red!80!black}{red} denotes failures. We observe that failures arise when the target gestures fall outside the distribution of the exploration data.} 
\label{fig:realworld}
\vskip -0.2in
\end{figure}
We validate MoDex in real-world in-hand manipulation and gesture generation. For in-hand manipulation, we use the Xhand (five-fingered dexterous hand) to manipulate a cube, orange, and cylinder at 10~Hz. Object states (position and orientation) are tracked via AprilTags using an Intel RealSense D415. The hand state is represented by joint positions, with 20 minutes of random exploration collected for pretraining. During evaluation, random $z$-axis rotations in $[0^{\circ}, 90^{\circ}]$ are used, with a success threshold of $3^{\circ}$.
For gesture generation, we use a RealSense D415 camera to track fingertip positions with CoTracker3~\cite{karaev24cotracker3}, collecting 10 minutes of random exploration. Instead of 3D positions, we directly use 2D pixel coordinates and provide pixel-based cost function examples to the LLM. See Appendix~\ref{sup:realworld} for details.

\begin{wraptable}[7]{r}{0.4\textwidth}
    \centering
    \vspace{-8pt}
    \captionsetup{font=small}
    \begin{tabular}{>{\centering\arraybackslash}l >{\centering\arraybackslash}p{0.18\textwidth}}
        \toprule[0.4mm]
        Task & Success Rate \\
        \midrule[0.2mm]
        Rotate Cube & 6/10  \\
        Rotate Orange & 2/10  \\
        Rotate Cylinder & 8/10 \\      
        \bottomrule
    \end{tabular}
    \vspace{-4pt}
    \caption{\textbf{Evaluation on in-hand rotation.}}
    \label{tab:real_inhand}
\end{wraptable}
As shown in Figure~\ref{fig:realworld}(a) and Table~\ref{tab:real_inhand}, MoDex achieves successful object rotations within 30 minutes of real-world rollouts, with performance varying by object. The orange is difficult to rotate beyond $30^{\circ}$ due to its size, while the cylinder rotates easily but is prone to slipping. The cube shows moderate difficulty, occasionally failing due to edge-related instabilities. MoDex also generates diverse hand gestures (Figure~\ref{fig:realworld}(b)), with failures typically arising when gestures fall outside the exploration data distribution, highlighting the importance of coverage during data collection.
\section{Conclusion and Limitations}
\label{sec:conclusion}

In this work, we present a neural network-based internal model for dexterous hand control in high-dimensional action space, achieving accurate fingertip reaching across four platforms when paired with a CEM planner. The model also supports data-efficient in-hand manipulation via an external model and gesture generation with an LLM in both simulation and real world. However, several limitations are worth further investigation. (1) The current approach depends on ground-truth states or external trackers, limiting real-world applicability. Replacing these with raw sensory inputs (e.g., point clouds or RGB) could enhance generality. (2) Additionally, while we use simple MLPs for dynamics modeling, more expressive architectures like diffusion models~\cite{ho2020denoising, song2020denoising, chi2023diffusion} may better handle the multimodal, stochastic nature of dexterous tasks.

\clearpage
\bibliography{main}

\begin{thebibliography}{37}
\providecommand{\natexlab}[1]{#1}
\providecommand{\url}[1]{\texttt{#1}}
\expandafter\ifx\csname urlstyle\endcsname\relax
  \providecommand{\doi}[1]{doi: #1}\else
  \providecommand{\doi}{doi: \begingroup \urlstyle{rm}\Url}\fi

\bibitem[Wang et~al.(2023)Wang, Zhang, Chen, Xu, Li, Liu, and Wang]{wang2023dexgraspnet}
R.~Wang, J.~Zhang, J.~Chen, Y.~Xu, P.~Li, T.~Liu, and H.~Wang.
\newblock {DexGraspNet}: A large-scale robotic dexterous grasp dataset for general objects based on simulation.
\newblock In \emph{Proceedings of IEEE International Conference on Robotics and Automation (ICRA)}, 2023.

\bibitem[Xu et~al.(2023)Xu, Wan, Zhang, Liu, Shan, Shen, Wang, Geng, Weng, Chen, Liu, Yi, and Wang]{xu2023unidexgrasp}
Y.~Xu, W.~Wan, J.~Zhang, H.~Liu, Z.~Shan, H.~Shen, R.~Wang, H.~Geng, Y.~Weng, J.~Chen, T.~Liu, L.~Yi, and H.~Wang.
\newblock {UniDexGrasp}: Universal robotic dexterous grasping via learning diverse proposal generation and goal-conditioned policy.
\newblock In \emph{Proceedings of IEEE/CVF Conference on Computer Vision and Pattern Recognition (CVPR)}, 2023.

\bibitem[Mandikal and Grauman(2021)]{mandikal2021learning}
P.~Mandikal and K.~Grauman.
\newblock Learning dexterous grasping with object-centric visual affordances.
\newblock In \emph{Proceedings of IEEE International Conference on Robotics and Automation (ICRA)}, 2021.

\bibitem[Handa et~al.(2023)Handa, Allshire, Makoviychuk, Petrenko, Singh, Liu, Makoviichuk, Van~Wyk, Zhurkevich, Sundaralingam, et~al.]{handa2023dextreme}
A.~Handa, A.~Allshire, V.~Makoviychuk, A.~Petrenko, R.~Singh, J.~Liu, D.~Makoviichuk, K.~Van~Wyk, A.~Zhurkevich, B.~Sundaralingam, et~al.
\newblock Dextreme: Transfer of agile in-hand manipulation from simulation to reality.
\newblock In \emph{Proceedings of IEEE International Conference on Robotics and Automation (ICRA)}, 2023.

\bibitem[Chen et~al.(2023)Chen, Tippur, Wu, Kumar, Adelson, and Agrawal]{doi:10.1126/scirobotics.adc9244}
T.~Chen, M.~Tippur, S.~Wu, V.~Kumar, E.~Adelson, and P.~Agrawal.
\newblock {Visual Dexterity}: In-hand reorientation of novel and complex object shapes.
\newblock \emph{Science Robotics}, 8\penalty0 (84):\penalty0 eadc9244, 2023.

\bibitem[Yin et~al.(2023)Yin, Huang, Qin, Chen, and Wang]{touch-dexterity}
Z.-H. Yin, B.~Huang, Y.~Qin, Q.~Chen, and X.~Wang.
\newblock Rotating without seeing: Towards in-hand dexterity through touch.
\newblock \emph{Proceedings of Robotics: Science and Systems (RSS)}, 2023.

\bibitem[Qi et~al.(2023)Qi, Yi, Suresh, Lambeta, Ma, Calandra, and Malik]{qi2023general}
H.~Qi, B.~Yi, S.~Suresh, M.~Lambeta, Y.~Ma, R.~Calandra, and J.~Malik.
\newblock General in-hand object rotation with vision and touch.
\newblock In \emph{Proceedings of Conference on Robot Learning (CoRL)}, 2023.

\bibitem[Wang et~al.(2022)Wang, Caggiano, Durandau, Sartori, and Kumar]{wang2022myosim}
H.~Wang, V.~Caggiano, G.~Durandau, M.~Sartori, and V.~Kumar.
\newblock {MyoSim}: Fast and physiologically realistic mujoco models for musculoskeletal and exoskeletal studies.
\newblock In \emph{Proceedings of IEEE International Conference on Robotics and Automation (ICRA)}, 2022.

\bibitem[Caggiano et~al.(2022{\natexlab{a}})Caggiano, Wang, Durandau, Sartori, and Kumar]{caggiano2022myosuite}
V.~Caggiano, H.~Wang, G.~Durandau, M.~Sartori, and V.~Kumar.
\newblock {MyoSuite}: A contact-rich simulation suite for musculoskeletal motor control.
\newblock In \emph{Proceedings of Learning for Dynamics and Control Conference (L4DC)}, 2022{\natexlab{a}}.

\bibitem[Caggiano et~al.(2022{\natexlab{b}})Caggiano, Durandau, Wang, Chiappa, Mathis, Tano, Patel, Pouget, Schumacher, Martius, Haeufle, Geng, An, Zhong, Ji, Chen, Dong, Yang, Siripurapu, Ferro~Diez, Kopp, Patil, Hochreiter, Tassa, Merel, Schultheis, Song, Sartori, and Kumar]{caggiano23myochallenge}
V.~Caggiano, G.~Durandau, H.~Wang, A.~Chiappa, A.~Mathis, P.~Tano, N.~Patel, A.~Pouget, P.~Schumacher, G.~Martius, D.~Haeufle, Y.~Geng, B.~An, Y.~Zhong, J.~Ji, Y.~Chen, H.~Dong, Y.~Yang, R.~Siripurapu, L.~E. Ferro~Diez, M.~Kopp, V.~Patil, S.~Hochreiter, Y.~Tassa, J.~Merel, R.~Schultheis, S.~Song, M.~Sartori, and V.~Kumar.
\newblock {MyoChallenge 2022}: Learning contact-rich manipulation using a musculoskeletal hand.
\newblock In \emph{Proceedings of the NeurIPS 2022 Competitions Track}, 2022{\natexlab{b}}.

\bibitem[Caggiano et~al.(2023)Caggiano, Dasari, and Kumar]{caggiano2023myodex}
V.~Caggiano, S.~Dasari, and V.~Kumar.
\newblock {MyoDex}: a generalizable prior for dexterous manipulation.
\newblock In \emph{Proceedings of International Conference on Machine Learning (ICML)}, 2023.

\bibitem[Grillner(1985)]{grillner1985neurobiological}
S.~Grillner.
\newblock Neurobiological bases of rhythmic motor acts in vertebrates.
\newblock \emph{Science}, 228\penalty0 (4696):\penalty0 143--149, 1985.

\bibitem[Zuo et~al.(2024)Zuo, He, Shao, and Sui]{zuo2024self}
C.~Zuo, K.~He, J.~Shao, and Y.~Sui.
\newblock Self model for embodied intelligence: Modeling full-body human musculoskeletal system and locomotion control with hierarchical low-dimensional representation.
\newblock In \emph{IEEE International Conference on Robotics and Automation (ICRA)}, 2024.

\bibitem[Berg et~al.(2023)Berg, Caggiano, and Kumar]{berg2023sar}
C.~Berg, V.~Caggiano, and V.~Kumar.
\newblock {SAR}: Generalization of physiological agility and dexterity via synergistic action representation.
\newblock \emph{Proceedings of Robotics: Science and Systems (RSS)}, 2023.

\bibitem[He et~al.(2024)He, Zuo, Ma, and Sui]{he2024dynsyn}
K.~He, C.~Zuo, C.~Ma, and Y.~Sui.
\newblock {DynSyn}: dynamical synergistic representation for efficient learning and control in overactuated embodied systems.
\newblock \emph{arXiv preprint arXiv:2407.11472}, 2024.

\bibitem[Kawato(1999)]{kawato1999internal}
M.~Kawato.
\newblock Internal models for motor control and trajectory planning.
\newblock \emph{Current Opinion in Neurobiology}, 9\penalty0 (6):\penalty0 718--727, 1999.

\bibitem[Tin and Poon(2005)]{tin2005internal}
C.~Tin and C.-S. Poon.
\newblock Internal models in sensorimotor integration: perspectives from adaptive control theory.
\newblock \emph{Journal of Neural Engineering}, 2\penalty0 (3):\penalty0 S147, 2005.

\bibitem[Egger et~al.(2019)Egger, Remington, Chang, and Jazayeri]{egger2019internal}
S.~W. Egger, E.~D. Remington, C.-J. Chang, and M.~Jazayeri.
\newblock Internal models of sensorimotor integration regulate cortical dynamics.
\newblock \emph{Nature Neuroscience}, 22\penalty0 (11):\penalty0 1871--1882, 2019.

\bibitem[Angelaki et~al.(2004)Angelaki, Shaikh, Green, and Dickman]{angelaki2004neurons}
D.~E. Angelaki, A.~G. Shaikh, A.~M. Green, and J.~D. Dickman.
\newblock Neurons compute internal models of the physical laws of motion.
\newblock \emph{Nature}, 430\penalty0 (6999):\penalty0 560--564, 2004.

\bibitem[Nagabandi et~al.(2018)Nagabandi, Kahn, Fearing, and Levine]{nagabandi2018neural}
A.~Nagabandi, G.~Kahn, R.~S. Fearing, and S.~Levine.
\newblock Neural network dynamics for model-based deep reinforcement learning with model-free fine-tuning.
\newblock In \emph{Proceedings of IEEE International Conference on Robotics and Automation (ICRA)}, 2018.

\bibitem[Shi et~al.(2024)Shi, Xu, Huang, Li, and Wu]{shi2024robocraft}
H.~Shi, H.~Xu, Z.~Huang, Y.~Li, and J.~Wu.
\newblock {RoboCraft}: Learning to see, simulate, and shape elasto-plastic objects in 3d with graph networks.
\newblock \emph{The International Journal of Robotics Research}, 43\penalty0 (4):\penalty0 533--549, 2024.

\bibitem[Rubinstein and Kroese(2004)]{rubinstein2004cross}
R.~Y. Rubinstein and D.~P. Kroese.
\newblock \emph{The cross-entropy method: a unified approach to combinatorial optimization, Monte-Carlo simulation, and machine learning}, volume 133.
\newblock Springer, 2004.

\bibitem[Francis and Wonham(1976)]{francis1976internal}
B.~A. Francis and W.~M. Wonham.
\newblock The internal model principle of control theory.
\newblock \emph{Automatica}, 12\penalty0 (5):\penalty0 457--465, 1976.

\bibitem[Wolpert et~al.(1995)Wolpert, Ghahramani, and Jordan]{wolpert1995internal}
D.~M. Wolpert, Z.~Ghahramani, and M.~I. Jordan.
\newblock An internal model for sensorimotor integration.
\newblock \emph{Science}, 269\penalty0 (5232):\penalty0 1880--1882, 1995.

\bibitem[Shadmehr and Mussa-Ivaldi(1994)]{shadmehr1994adaptive}
R.~Shadmehr and F.~A. Mussa-Ivaldi.
\newblock Adaptive representation of dynamics during learning of a motor task.
\newblock \emph{Journal of Neuroscience}, 14\penalty0 (5), 1994.

\bibitem[Yuan et~al.(2024)Yuan, Che, Qin, Huang, Yin, Lee, Wu, Lim, and Wang]{yuan2023robot}
Y.~Yuan, H.~Che, Y.~Qin, B.~Huang, Z.-H. Yin, K.-W. Lee, Y.~Wu, S.-C. Lim, and X.~Wang.
\newblock {Robot Synesthesia}: In-hand manipulation with visuotactile sensing.
\newblock In \emph{Proceedings of IEEE International Conference on Robotics and Automation (ICRA)}, 2024.

\bibitem[Nagabandi et~al.(2020)Nagabandi, Konolige, Levine, and Kumar]{nagabandi2020deep}
A.~Nagabandi, K.~Konolige, S.~Levine, and V.~Kumar.
\newblock Deep dynamics models for learning dexterous manipulation.
\newblock In \emph{Proceedings of Conference on Robot Learning (CoRL)}, 2020.

\bibitem[Ito(2008)]{ito2008control}
M.~Ito.
\newblock Control of mental activities by internal models in the cerebellum.
\newblock \emph{Nature Reviews Neuroscience}, 9\penalty0 (4):\penalty0 304--313, 2008.

\bibitem[Woodworth(1899)]{woodworth1899accuracy}
R.~S. Woodworth.
\newblock Accuracy of voluntary movement.
\newblock \emph{The Psychological Review: Monograph Supplements}, 3\penalty0 (3):\penalty0 i, 1899.

\bibitem[Elliott et~al.(2001)Elliott, Helsen, and Chua]{elliott2001century}
D.~Elliott, W.~F. Helsen, and R.~Chua.
\newblock A century later: Woodworth's (1899) two-component model of goal-directed aiming.
\newblock \emph{Psychological bulletin}, 127\penalty0 (3):\penalty0 342, 2001.

\bibitem[Huang et~al.(2023)Huang, Wang, Zhang, Li, Wu, and Fei-Fei]{huang2023voxposer}
W.~Huang, C.~Wang, R.~Zhang, Y.~Li, J.~Wu, and L.~Fei-Fei.
\newblock {VoxPoser}: Composable {3D} value maps for robotic manipulation with language models.
\newblock In \emph{Proceedings of Conference on Robot Learning (CoRL)}, 2023.

\bibitem[Makoviychuk et~al.(2021)Makoviychuk, Wawrzyniak, Guo, Lu, Storey, Macklin, Hoeller, Rudin, Allshire, Handa, and State]{makoviychuk2021isaac}
V.~Makoviychuk, L.~Wawrzyniak, Y.~Guo, M.~Lu, K.~Storey, M.~Macklin, D.~Hoeller, N.~Rudin, A.~Allshire, A.~Handa, and G.~State.
\newblock {Isaac Gym}: High performance {GPU} based physics simulation for robot learning.
\newblock In \emph{Proceedings of Neural Information Processing Systems Track on Datasets and Benchmarks (NeurIPS)}, 2021.

\bibitem[Haarnoja et~al.(2018)Haarnoja, Zhou, Abbeel, and Levine]{haarnoja2018soft}
T.~Haarnoja, A.~Zhou, P.~Abbeel, and S.~Levine.
\newblock Soft actor-critic: Off-policy maximum entropy deep reinforcement learning with a stochastic actor.
\newblock In \emph{Proceedings of International Conference on Machine Learning (ICML)}, 2018.

\bibitem[Karaev et~al.(2023)Karaev, Makarov, Wang, Neverova, Vedaldi, and Rupprecht]{karaev24cotracker3}
N.~Karaev, I.~Makarov, J.~Wang, N.~Neverova, A.~Vedaldi, and C.~Rupprecht.
\newblock {CoTracker3}: Simpler and better point tracking by pseudo-labelling real videos.
\newblock \emph{arXiv preprint arXiv:2410.11831}, 2023.

\bibitem[Ho et~al.(2020)Ho, Jain, and Abbeel]{ho2020denoising}
J.~Ho, A.~Jain, and P.~Abbeel.
\newblock Denoising diffusion probabilistic models.
\newblock \emph{Advances in neural information processing systems (NeurIPS)}, 2020.

\bibitem[Song et~al.(2021)Song, Meng, and Ermon]{song2020denoising}
J.~Song, C.~Meng, and S.~Ermon.
\newblock Denoising diffusion implicit models.
\newblock In \emph{Proceedings of International Conference on Learning Representations (ICLR)}, 2021.

\bibitem[Chi et~al.(2023)Chi, Feng, Du, Xu, Cousineau, Burchfiel, and Song]{chi2023diffusion}
C.~Chi, S.~Feng, Y.~Du, Z.~Xu, E.~Cousineau, B.~C. Burchfiel, and S.~Song.
\newblock {Diffusion Policy}: Visuomotor policy learning via action diffusion.
\newblock In \emph{Proceedings of Robotics: Science and Systems (RSS)}, 2023.

\end{thebibliography}

\clearpage
\newpage
\appendix
\section*{Appendix}

\startcontents[appendices]
\printcontents[appendices]{l}{0}{\setcounter{tocdepth}{2}}

\clearpage

\section{Analysis on Synergy}
\label{sup:syn}
To analyze the variance of motor synergies across different tasks, we conduct an experiment comparing the synergies of ``Chopstick Using" and ``Cube Rotating." A subject wears a data glove while performing each task, and 1,000 data samples are collected per case. Principal Component Analysis (PCA) and Independent Component Analysis (ICA) are applied to extract the underlying synergies ~\cite{berg2023sar}.

\begin{figure}[h]
\centering
\includegraphics[width=0.8\textwidth]{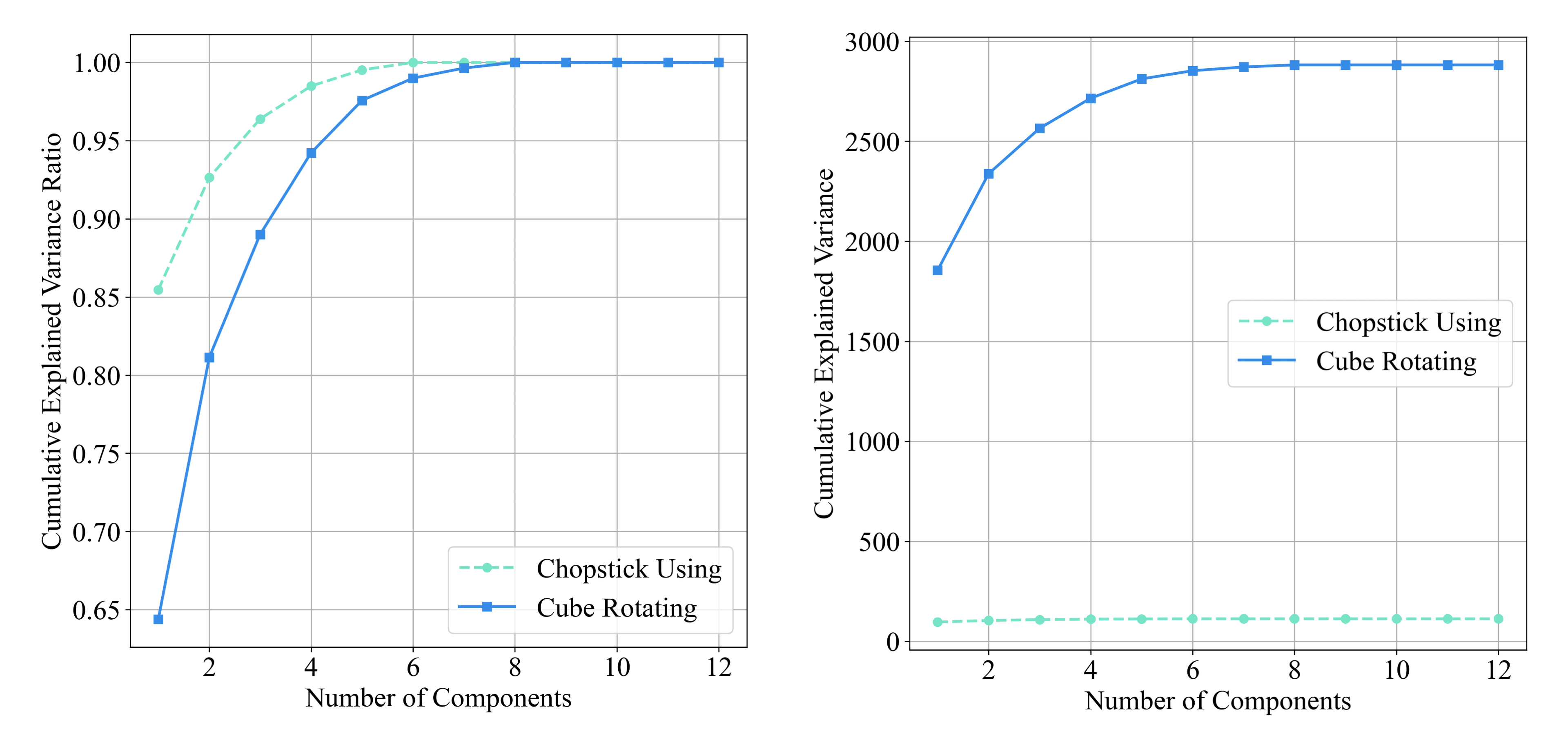}
\caption{\textbf{Explained variance ratio and explained variance.}} 
\label{fig:ev}
\vskip -0.2in
\end{figure}

Figure~\ref{fig:ev} presents the cumulative explained variance and cumulative explained variance ratio. The cumulative explained variance ratio demonstrates that ``Chopstick Using" reaches near 1.0 more rapidly than ``Cube Rotating," suggesting that fewer principal components are required to capture its variance. In contrast, ``Cube Rotating" exhibits a significantly higher cumulative explained variance, indicating greater movement complexity and variability. These findings suggest that the number of synergies varies across tasks.

\begin{figure}[h]
\centering
\includegraphics[width=0.8\textwidth]{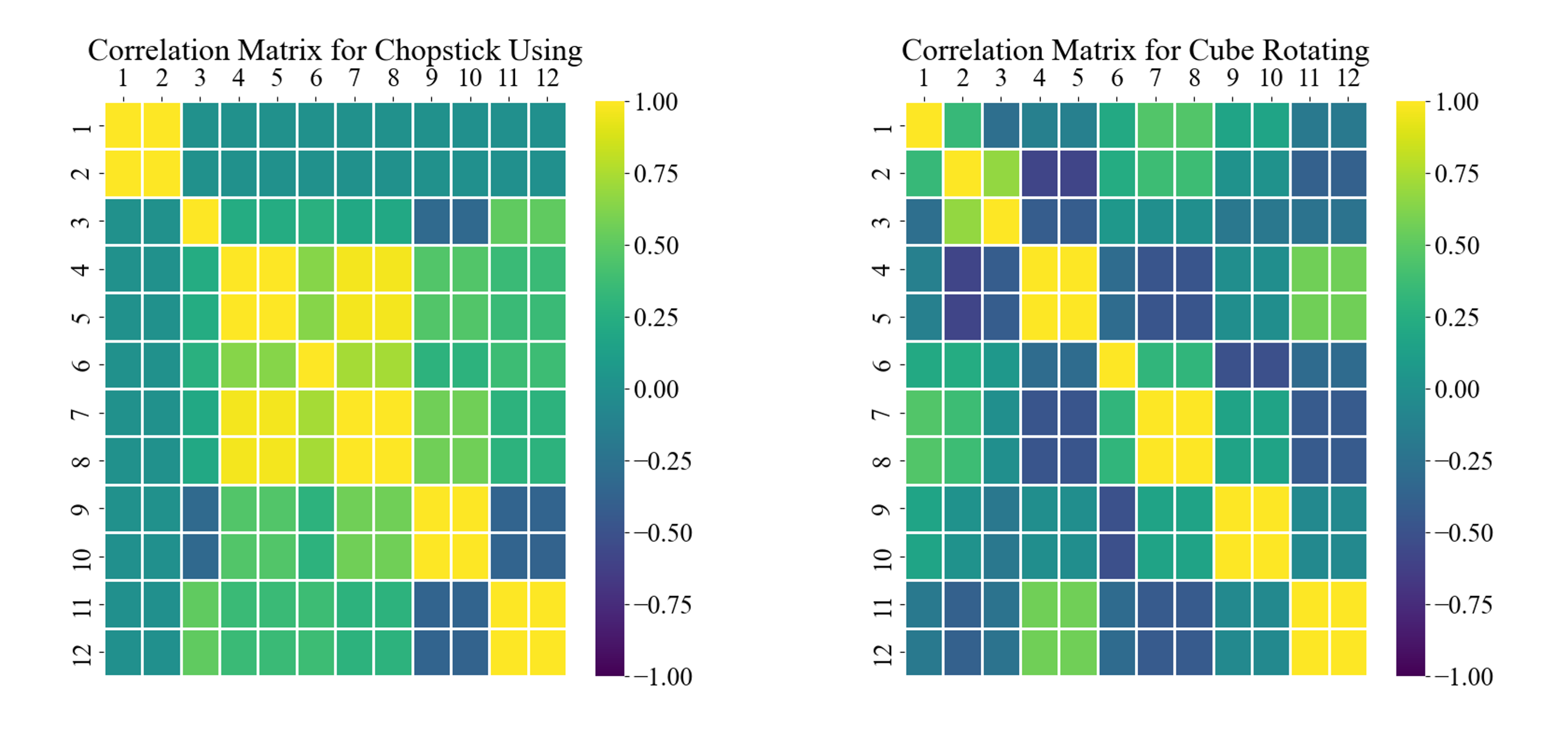}
\caption{\textbf{Correlation matrices.}} 
\label{fig:cormat}
\vskip -0.2in
\end{figure}

Additionally, the correlation matrix of action dimensions, visualized in Figure~\ref{fig:cormat}, reveals distinct correlation patterns between the two tasks. This further supports that the underlying synergies differ between ``Chopstick Using" and ``Cube Rotating," addressing the task-dependent nature of motor synergies.

\begin{figure}[h]
\centering
\includegraphics[width=1\textwidth]{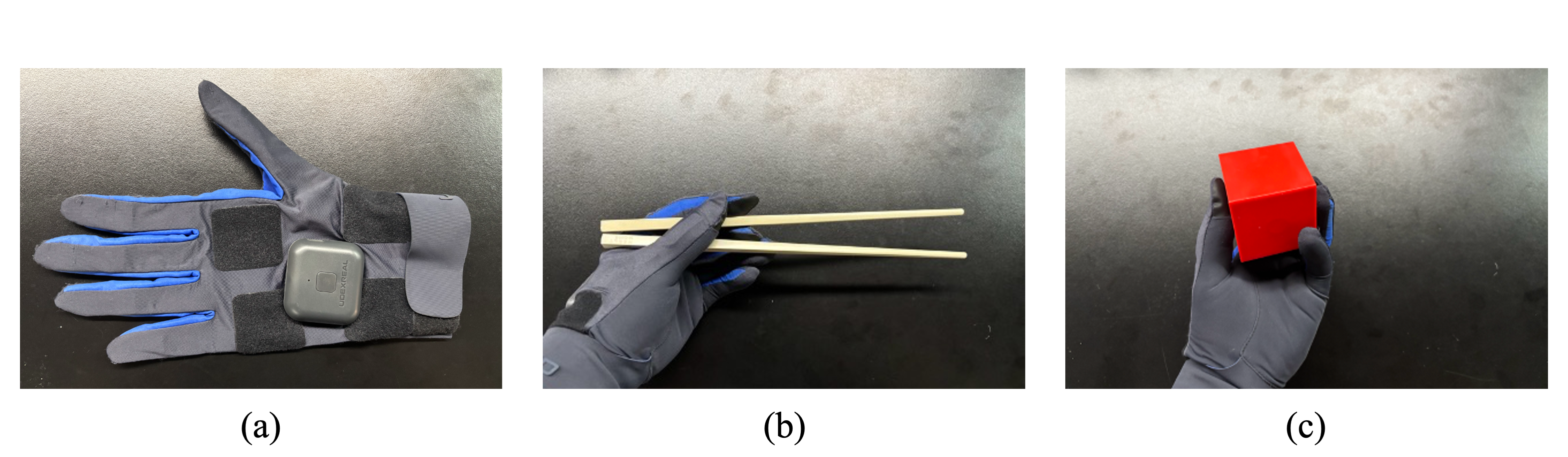}
\caption{\textbf{Synergy analysis experiment.}} 
\label{fig:syn}
\vskip -0.2in
\end{figure}

\section{Dexterous Hands in Simulation}
\label{sup:hands}
We investigate simulation models of four different dexterous hands, which cover a range of driving mechanisms (joint-driven and tendon-driven), actuation types (fully-actuated, under-actuated, and over-actuated), and structural complexities, reflected in their number of fingers (NoFs), degrees of freedom (DoFs), and action dimensions.

\begin{table}[h]
\centering
\caption{\textbf{Dexterous hands.}}
\begin{tabular}{lllccc}
\toprule
\textbf{Name}        & \textbf{Driven}      & \textbf{Actuation}     & \textbf{NoF} & \textbf{DoF} & \textbf{Dimension} \\ 
\midrule
Robotiq   & Joint-driven         & Fully-actuated         & 3             & 11            & 11                 \\ 
Allegro     & Joint-driven         & Fully-actuated         & 4             & 16            & 16                 \\ 
Shadowhand  & Joint-driven         & Under-actuated         & 5             & 24            & 20                 \\ 
Myohand   & Tendon-driven        & Over-actuated          & 5             & 23            & 39                 \\ 
\bottomrule
\end{tabular}
\label{table:hands}
\end{table}

\section{Bidirectional Planning Details}
\label{sup:biplan}

For sequential problems, we generate an action distribution at each step using the inverse model. We expand the distribution over the planning horizon $T_0$ and then sample actions. For $N_{cem}$ iterations, we predict action trajectories using the forward model, compute trajectory costs, and select the top $\rho$ fraction as elites. The action distribution is then updated based on the mean and variance of the elite trajectories. Finally, we obtain an optimized action sequence and execute the first action. At the next step, we replan using the updated state.

\begin{algorithm}[h]
\caption{Bidirectional Planning with MPC}
\label{alg:bidirectional_planning}
\begin{algorithmic}[1]
\Require Current state $\boldsymbol{s}_t$, target state $\boldsymbol{s}_T$, inverse model $g_{\phi}$, forward model $f_{\theta}$, planning horizon $T_0$, number of CEM iterations $N_\text{cem}$, number of samples $N_s$, number of elites $N_\text{elites}$, filtering coefficient (moving average coefficient) $\beta$
\While{task is not complete}
    \State \textbf{Initialize:} Generate initial action distribution:
    \State \quad $\boldsymbol{\hat{a}} \sim \mathcal{N}(g_{\phi}(\boldsymbol{s}_t,\boldsymbol{s}_T), \Sigma)$
    \State Expand the distribution to planning horizon $T_0$: $\{\boldsymbol{\hat{a}}\}_{i=t}^{t+T_0-1}$
    \For{iteration $i = 1$ to $N_\text{cem}$}
        \State Sample $N_s$ candidate action sequences from the current distribution
        \State \textbf{Initialize:} $\hat{\boldsymbol{s}}_t = {\boldsymbol{s}}_t$
        \For{each candidate action sequence $\{\boldsymbol{a}_t\}_{t=0}^{T_0-1}$}
            \State Predict trajectory using forward model:
            \State \quad $\hat{\boldsymbol{s}}_{i+1} = f_{\theta}(\hat{\boldsymbol{s}}_i, \boldsymbol{a}_i), \quad \forall i \in [t, t+T_0-1]$
            \State Compute trajectory cost $J$ based on task objective
        \EndFor
        \State Select top $N_\text{elites}$ action sequences with lowest costs (elites)
        \State Calculate the mean and variance of elite trajectories
        \State Update action distribution using moving average method
    \EndFor
    \State Execute first action $\boldsymbol{a}_t$
    \State Update state to $\boldsymbol{s}_{t+1}$
\EndWhile
\end{algorithmic}
\end{algorithm}

\section{Analysis on Factorized Dynamics}
\label{sup:facdyn}

A complete system model requires learning an input-output mapping of dimension $(H+O+K) \times (H+O)$ where $H$, $O$, and $K$ are the dimensions of hand state, object state, and action space, respectively.
When the action dimension increases, the dimension of the input-output mapping scales proportionally by a factor of $(H+O)$, leading to higher model complexity and data consumption. Instead, factorized dynamics reduces this to $(H+O) \times (H+O)$ by freezing the internal model and only train the external model, effectively eliminating direct dependence on action dimension.

\section{Quasi-Static and Sequential Settings}
\label{sup:setting}
We distinguish dexterous control tasks into two settings. \textbf{Quasi-static setting} focuses solely on the final outcome and is formulated as a single-step task. In contrast, \textbf{sequential setting} considers intermediate states and is formulated as a multi-step task.

In the field of dexterous control, sequential tasks like in-hand manipulation are a major area of focus. However, non-sequential tasks, where the static result is more important than the dynamic process, also play a crucial role. Examples include gesture generation and dexterous grasping. Therefore, we include both settings in our experiment.

\begin{figure}[h]
\centering
\vspace{-0.3cm}
\includegraphics[width=0.6\textwidth]{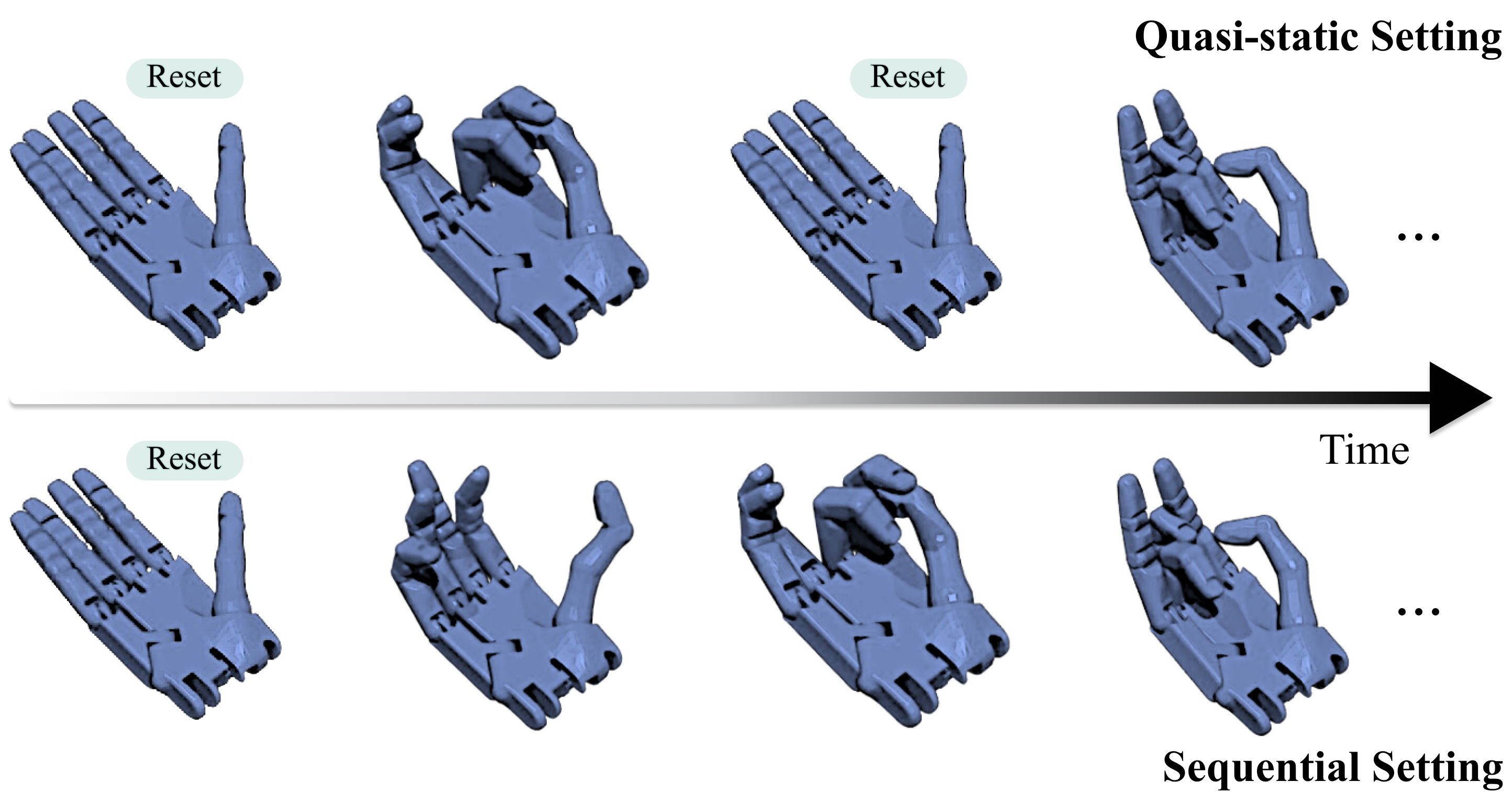}
\caption{\textbf{Quasi-static setting v.s. sequential setting.}} 
\label{fig:setting}
\vskip -0.2in
\end{figure}

\section{Fingertip Reach Experiment Details}
\subsection{Task Details}
We list the task details in Table~\ref{tab:task_info}.
\begin{table}[h]
    \centering
    \caption{\textbf{Task settings for different dexterous hands.}}
    \begin{tabular}{lcccc}
        \toprule
        \textbf{Task Information} & \textbf{Robotiq} & \textbf{Allegro} & \textbf{Shadowhand} & \textbf{Myohand} \\
        \midrule
        State dim & $9$ & $12$ & $15$ & $15$ \\
        Action dim & $11$ & $16$ & $20$ & $39$ \\
        Quasi-static skipped frames  & $150$ & $50$ & $200$ & $100$ \\
        Sequential skipped frames  & $10$ & $10$ & $10$ & $5$ \\
        Successful threshold (cm)  & $1.50$  & $1.50$ & $1.50$ & $1.25$ \\
        \bottomrule
    \end{tabular}
    \label{tab:task_info}
\end{table}

\subsection{Hyperparameters}
We list the hyperparameters of fingertip reach experiment in Table~\ref{tab:reach_hyperparams}.
\begin{table}[h]
    \centering
    \caption{\textbf{Hyperparameter settings.}}
    \begin{tabular}{lcc}
        \toprule
        \textbf{Hyperparameters} & \textbf{Quasi-static} & \textbf{Sequential} \\
        \midrule
        \multicolumn{3}{l}{\textbf{Forward model}} \\ 
        \cmidrule{1-3}
        Prediction horizon $S$ & $1$ & $10$ \\
        Discount factor $\alpha$  & \textendash & $0.95$ \\
        Optimizer  & Adam & Adam \\
        Learning rate &  $1e^{-4}$ & $1e^{-4}$ \\
        \midrule
        \multicolumn{3}{l}{\textbf{Inverse model}} \\ 
        \cmidrule{1-3}
        Start timestamp shift $t_0$ & $1$ & $1$ \\
        Optimizer  & Adam  & Adam \\
        Learning rate &  $1e^{-4}$ & $1e^{-4}$ \\
        \midrule
        \multicolumn{3}{l}{\textbf{Planning}} \\ 
        \cmidrule{1-3}
        Planning horizon $T_0$ & $1$ & $3$ \\ 
        Number of CEM iterations $N_{cem}$  & $5$ & $3$ \\
        Number of samples $N_s$  & $400$ & $600$ \\
        Number of elite $N_\text{elites}$  & $20$ &  $20$\\
        Filtering coefficient $\beta$ & $0.1$ & $0.2$ \\
        \bottomrule
    \end{tabular}
    \label{tab:reach_hyperparams}
\end{table}

\subsection{Implementation Details}
For model-free methods, policies are learned online. Model-based methods, however, involve a training phase where data is collected using a random policy to train a forward model of the hand. In MoDex, the same data is also used to train an inverse model. During evaluation, the forward model is frozen, and different planning methods are applied.

\subsection{Additional Results}
\label{sup:reachresults}

\begin{figure}[h]
\centering
\includegraphics[width=0.75\textwidth]{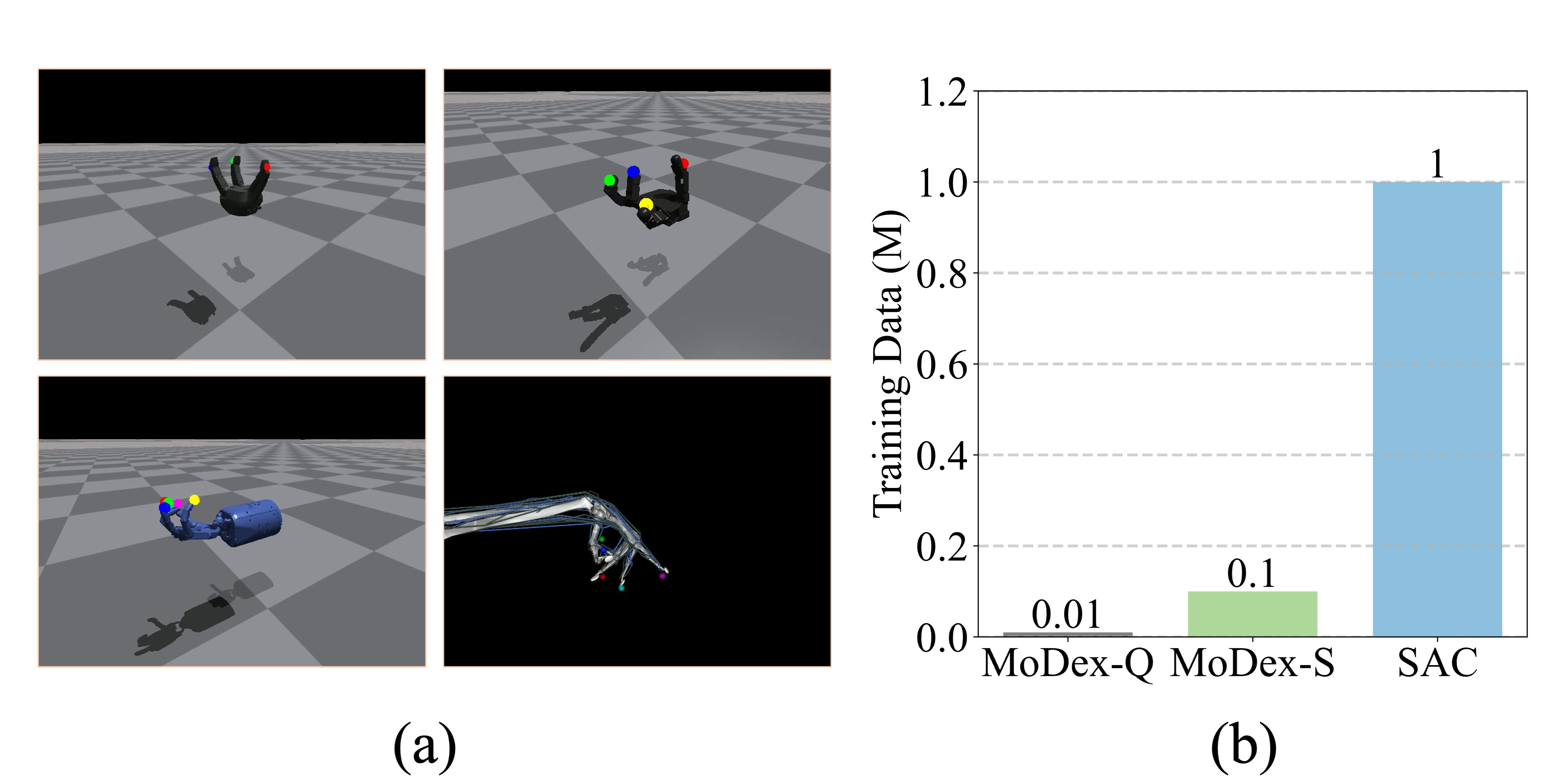}
\caption{\textbf{(a)} Fingertip reach environment. \textbf{(b)} Data consumption of MoDex (and other model-based methods) and SAC (and SAR).} 
\label{fig:reach}
\vskip -0.2in
\end{figure}

We provides the snapshots of fingertip reach task in Figure~\ref{fig:reach}(a) and visualize the data consumption of model-based methods and model-free methods in Figure~\ref{fig:reach}(b).

\begin{figure}[h]
\centering
\includegraphics[width=0.75\textwidth]{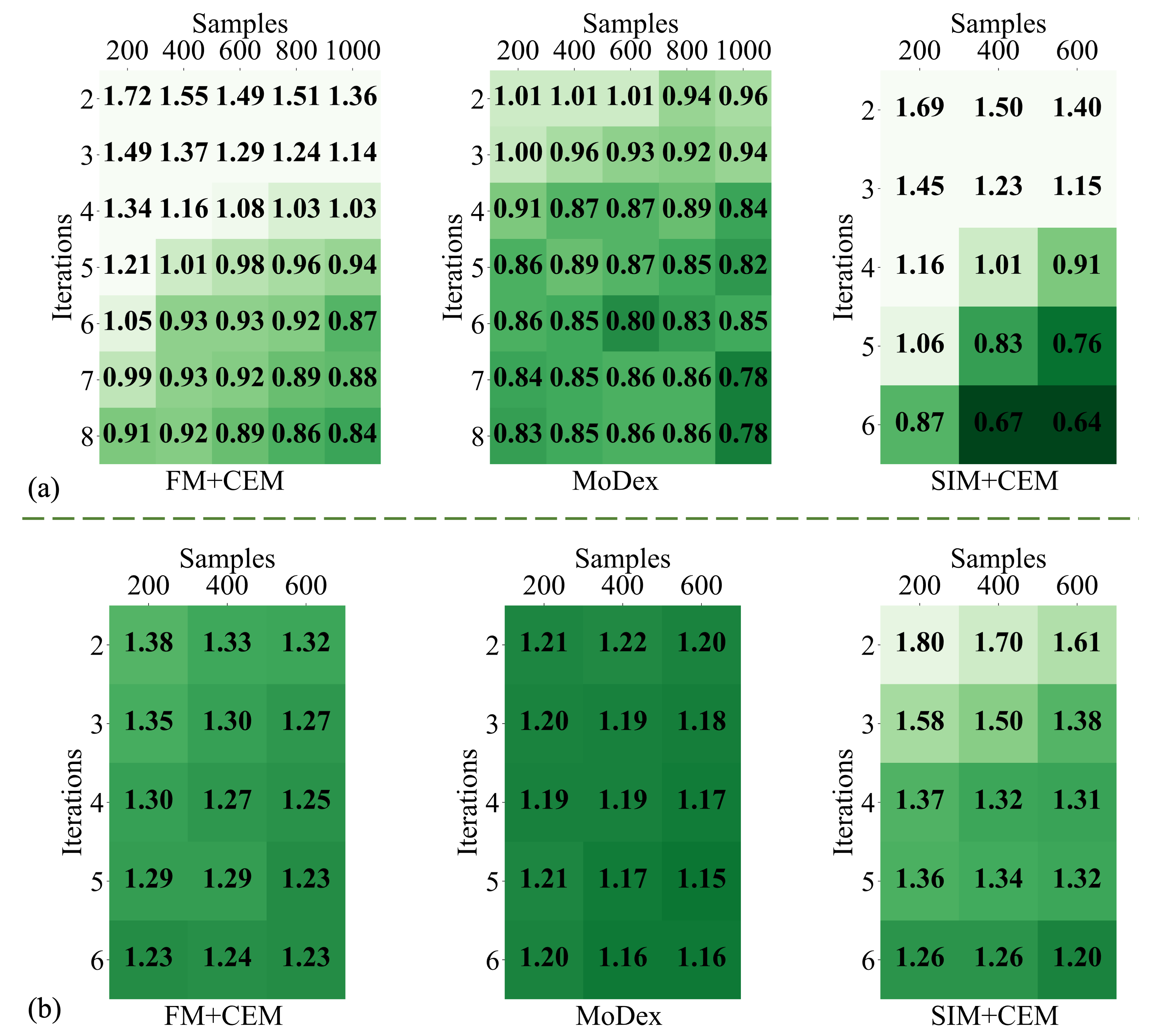}
\caption{Additional reach error results on bidirectional planning and CEM using Myohand in \textbf{(a)} quasi-static setting and \textbf{(b)} sequential setting.} 
\label{fig:add1}
\end{figure}

\begin{figure}[h]
\centering
\includegraphics[width=0.7\textwidth]{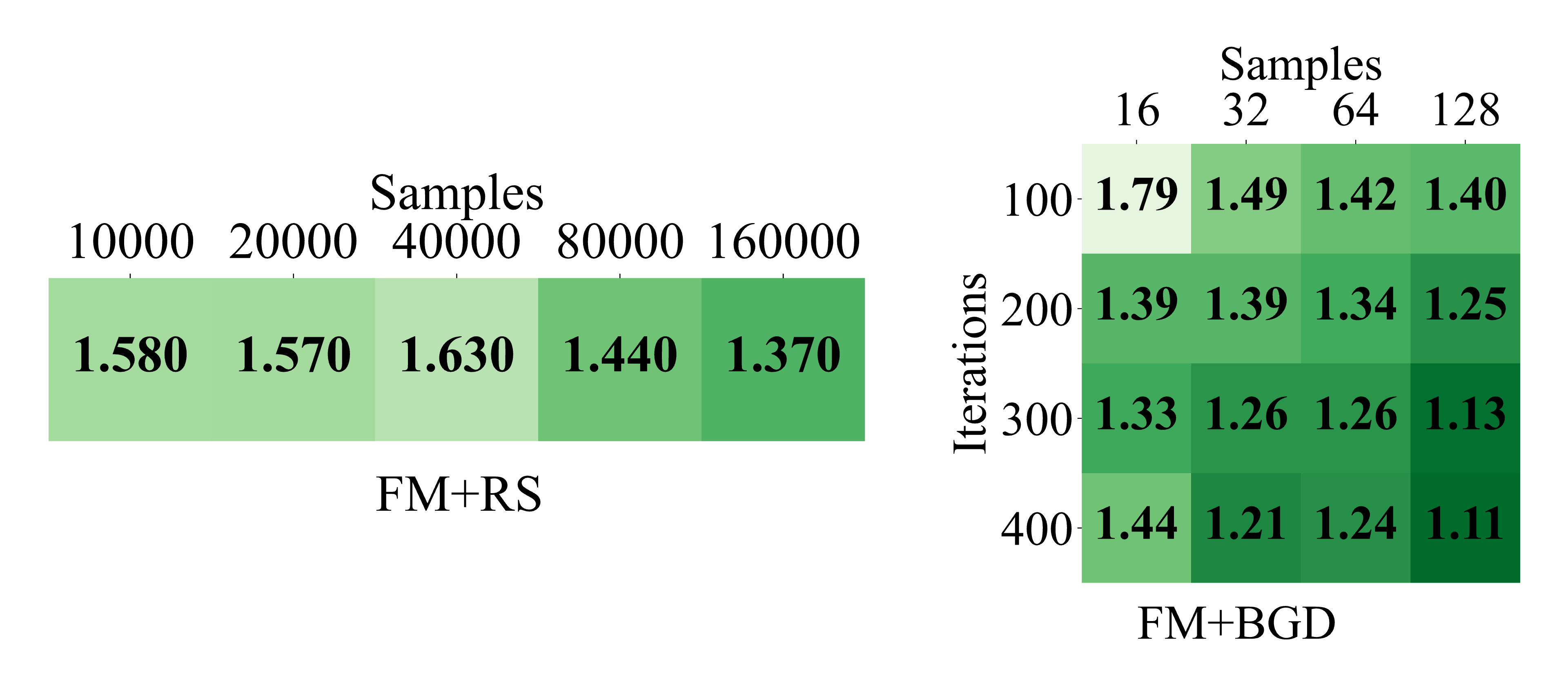}
\caption{Additional reach error results on RS and BGD using Myohand in quasi-static setting.} 
\label{fig:add2}
\end{figure}

\textbf{More results on model-based methods.} In Figure~\ref{fig:add1}, we introduce SIM+CEM as an additional baseline, which directly employs the ground-truth simulator as the forward model for planning. As shown in Figure~\ref{fig:add1}, MoDex achieves the same reach error with fewer iterations and samples compared to FM+CEM and even SIM+CEM. SIM+CEM achieves lower reach error due to its powerful forward model; however, its planning efficiency is limited. These results demonstrate the effectiveness of bidirectional planning and the importance of efficiency in high-dimensional space. In Figure~\ref{fig:add2}, we further analyze FM+RS and FM+BGD. FM+RS demonstrates low task completion rate and poor planning efficiency due to its reliance on random sampling. Although gradient-based methods achieve state-of-the-art performance in \cite{shi2024robocraft} , they face challenges in high-dimensional action space under constrained planning budgets.

\textbf{Adaptability to actuator changes.} 
We model hand-related changes, including actuator failures (by setting the broken actuator's command to zero) and muscle fatigue (by scaling commands by a factor), to assess the forward model's adaptability. For each scenario, we collect 1,000 samples for fine-tuning and 500 for evaluation. As shown in Figure~\ref{fig:ablation}(b), the NN-based model successfully adapts to these changes while maintaining prediction accuracy. Moreover, higher fatigue levels increase adaptation time, whereas the effect of actuator failures varies, likely due to structural difference of each broken actuator.

\begin{figure}[h]
\centering
\includegraphics[width=0.8\textwidth]{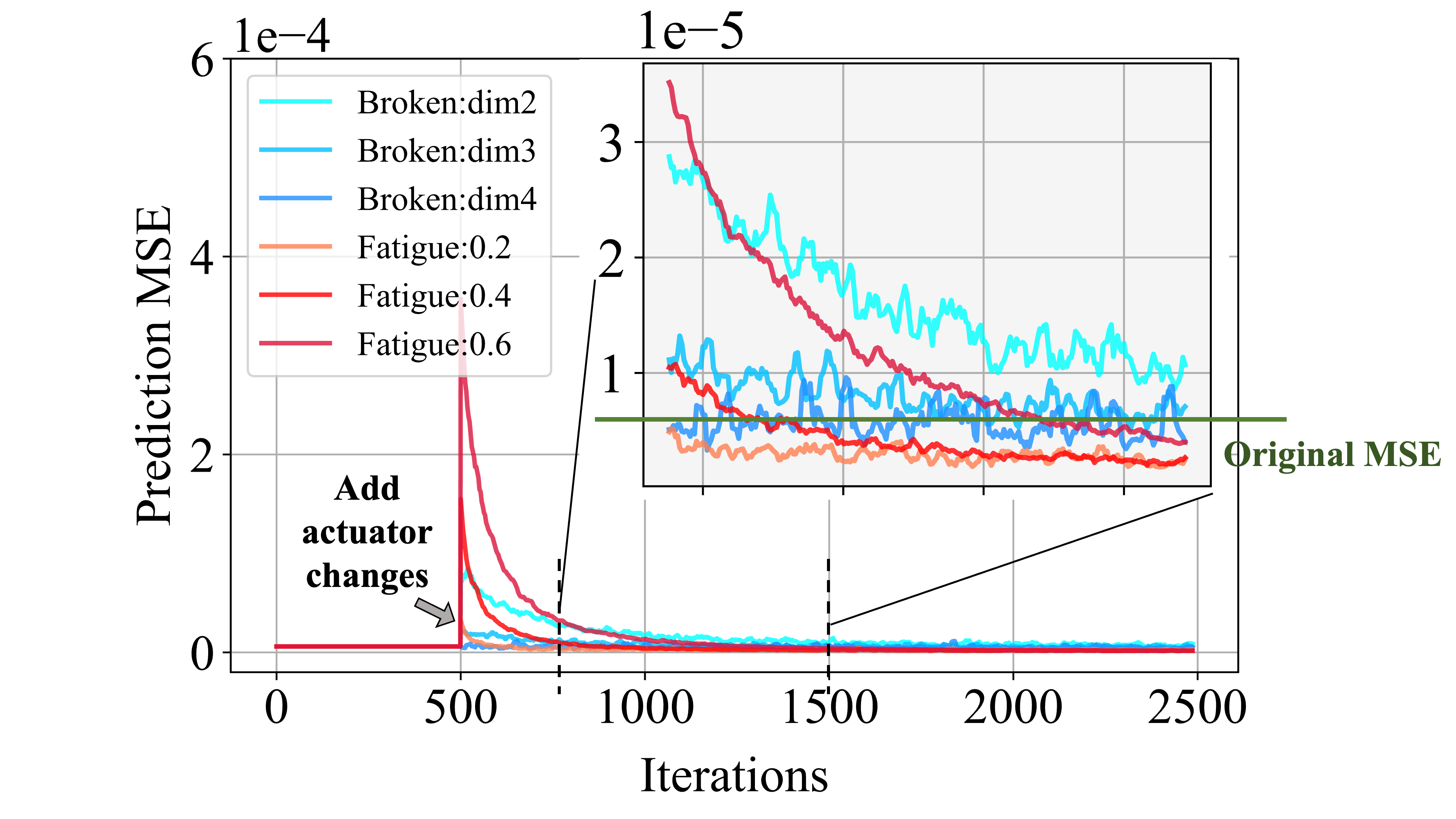}
\caption{\textbf{Adaptation of forward model to different actuator changes.}} 
\label{fig:adaptation}
\end{figure}

\section{In-hand Manipulation Details}
\subsection{Task Description}

\textbf{Reorient target2.} A Myohand reorients a cube to one of two goal orientations, $\pm\frac{\pi}{2}$ around $z$ axis. Each episode starts with a randomly initialized goal and cube position. The task is completed when $\text{cos}(\text{rot}_{\text{cube}},  \text{rot}_{\text{goal}})>0.95$ and $|\text{pos}_{\text{cube}} - \text{pos}_{\text{goal}}|<0.075$.

\textbf{Reorient 8object.} A Myohand manipulates one of four geometries—ellipsoid, cuboid, cylinder, or capsule—each with two scaled variants. The object pose and goal orientation are randomly initialized at the start of each episode. The object is randomly selected from the 8 geometry variants.

\textbf{Reorient 100object.}  A Myohand reorients objects sampled from 100 geometric variants (25 variants for each geometry). The object pose and goal orientation are randomly initialized at the start of each episode. The object is randomly selected from the 100 geometry variants.

\begin{figure}[h]
\centering
\includegraphics[width=0.7\textwidth]{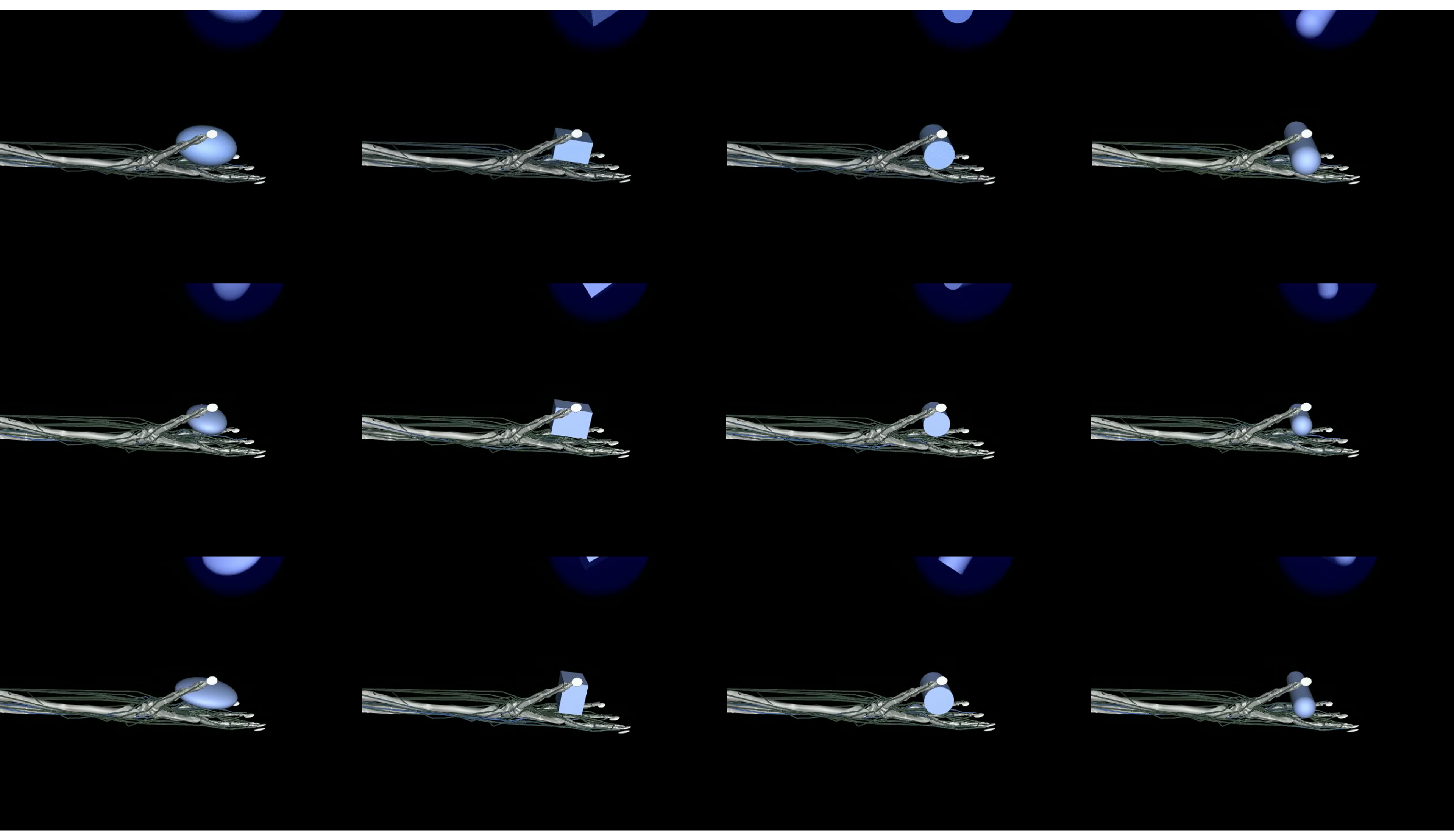}
\caption{\textbf{Snapshots of different geometry variants.}} 
\vskip -0.2in
\label{fig:inhand_envs}
\end{figure}

We sample several geometry variants in Figure~\ref{fig:inhand_envs}.

\subsection{Reward/Cost Function}

The reward (or negative cost) function used is defined as:
\begin{equation}
R := -\lambda_1 | \text{pos}_{\text{object}} - \text{pos}_{\text{goal}} |_2 + \lambda_2 \cos(\text{rot}_{\text{object}}, \text{rot}_{\text{goal}}) - \lambda_3 \cdot \mathbb{1}_{{\text{dropped}}},
\end{equation}
where $\lambda_1 = 1$, $\lambda_2 = 1$, and $\lambda_3 = 5$. The first term penalizes the Euclidean distance between the object and goal positions, the second term rewards rotational alignment via cosine similarity, and the final term imposes a penalty if the object is dropped, indicated by the indicator function $\mathbb{1}_{{\text{dropped}}}$.

\subsection{Hyperparameter}
We list the hyperparameters of in-hand manipulation in Table~\ref{tab:inhand_hyperparams}.

\begin{table}[h]
    \centering
    \caption{\textbf{Hyperparameter settings for in-hand manipulation.}}
    \begin{tabular}{lc}
        \toprule
        \textbf{Hyperparameters} & \textbf{Value} \\
        \midrule
        \multicolumn{2}{l}{\textbf{Rollout}} \\ 
        \cmidrule{1-2}
        Number of iterations     &  200 \\
        Number of rollouts per iteration & 40\\
        \midrule
        \multicolumn{2}{l}{\textbf{Planning}} \\ 
        \cmidrule{1-2}
        Planning horizon  & $7$ \\
        Number of Iterations & $3$ \\
        Number of samples & $1000$ \\
        Number of elites  &  20 \\
        Filtering coefficient $\beta$ & $0.2$ \\
        \midrule
        \multicolumn{2}{l}{\textbf{Factorize dynamics}} \\ 
        \cmidrule{1-2}
        Prediction horizon & $10$ \\
        Discount factor  & $0.95$ \\
        Optimizer & Adam \\
        Learning rate & $1e^{-4}$ \\
        \bottomrule
    \end{tabular}
    \label{tab:inhand_hyperparams}
\end{table}

\section{Gesture Generation Details}
\label{sup:gesture}

We provide the system prompt and several generated cost functions for Allegro as follows.

\begin{tcolorbox}
[breakable,title=\textbf{Prompt of gesture generation}]

You are acquired to generate code to control a robotic hand to make gestures, here are some examples:\\
\\
def gesture\_thumbup\_allegro(fingertip\_position):\\
    ''' \\
    num\_envs (1) x finger\_num x 3\\
    for allegro\\
    indices: indexs: 0-thumb, 1-ring finger\&little finger, 2-middle finger, 3-index finger\\
    '''\\
    \# loss that forces the thumb to be straight\\
    loss1 = -fingertip\_position[:,0,1]\\
    \# loss that forces the index finger, middle finger and ring finger to curl towards palm\\
    loss2 = fingertip\_position[:,1:, 2].mean(-1)\\
    \# combine the losses for all fingers\\
    loss = 1. * loss1 + 1. * loss2\\
    return loss\\

def gesture\_ok\_allegro(fingertip\_position):\\
    ''' \\
    num\_envs (1) x finger\_num x 3\\
    for allegro\\
    indices: 0-thumb, 1-ring finger\&little finger, 2-middle finger, 3-index finger\\
    '''\\
    \# loss that forces the tips of thumb and index finger to be close together\\
    loss1 = torch.norm(fingertip\_position[:,0] - fingertip\_position[:,3], dim=-1)\\
    \# loss that forces the middle finger and ring finger to be straight\\
    loss2 =  -fingertip\_position[:,1:3, 2].mean(dim=-1)\\
    \# combine the losses for all fingers\\
    loss = 1. * loss1 + 1. * loss2\\
    return loss\\

Refer to the examples and generate gestures according to the users' request.
\end{tcolorbox}

\begin{tcolorbox}
[breakable,title=\textbf{Generated cost functions}]

def gesture\_scissors\_allegro(fingertip\_position):\\
    '''\\ 
    num\_envs (1) x finger\_num x 3\\
    for allegro\\
    indices: 0-thumb, 1-ring finger\&little finger, 2-middle finger, 3-index finger\\
    '''\\
    \# loss that forces the tips of thumb, ring finger and little finger to be close together \\
    loss1 = torch.norm(fingertip\_position[:,0] - fingertip\_position[:,1], dim=-1)\\
    \# loss that forces the index finger and middle finger to be straight\\
    loss2 =  -fingertip\_position[:,2:4, 2].mean(dim=-1)\\
    \# combine the losses for all fingers \\
    loss = 1. * loss1 + 1. * loss2 \\
    return loss\\

def gesture\_rockandroll\_allegro(fingertip\_position):\\
    ''' \\
    num\_envs (1) x finger\_num x 3\\
    for allegro\\
    indices: 0-thumb, 1-ring finger\&little finger, 2-middle finger, 3-index finger\\
    '''\\
    \# loss that forces the tips of thumb and middle finger to be close together\\
    loss1 = torch.norm(fingertip\_position[:,0] - fingertip\_position[:,2], dim=-1)\\
    \# loss that forces the index finger, ring finger and little finger to be straight\\
    loss2 = - (fingertip\_position[:,1, 2] + fingertip\_position[:,3,2])/2\\
    \# combine the losses for all fingers\\
    loss = loss1 + 1. * loss2\\
    return loss\\

def gesture\_call\_allegro(fingertip\_position):\\
    ''' \\
    num\_envs (1) x finger\_num x 3\\
    for allegro\\
    indices: 0-thumb, 1-ring finger\&little finger, 2-middle finger, 3-index finger\\
    '''\\
    \# loss that forces the tips of thumb to be straight\\
    loss1 = -fingertip\_position[:,0,1]\\
    \# loss that forces the little finger to be straight\\
    loss2 = fingertip\_position[:,1, 0]\\
    \# loss that forces the index finger and middle finger to curl towards palm\\
    loss3 = fingertip\_position[:,2:, 2].mean(-1)\\
    \# combine the losses for all fingers\\
    loss = 1. * loss1 + 1. * loss2 + 1. * loss3\\
    return loss\\

    
\end{tcolorbox}

\clearpage

\section{Real-World Deployment Details}
\label{sup:realworld}

\begin{figure}[h]
\centering
\includegraphics[width=0.9\textwidth]{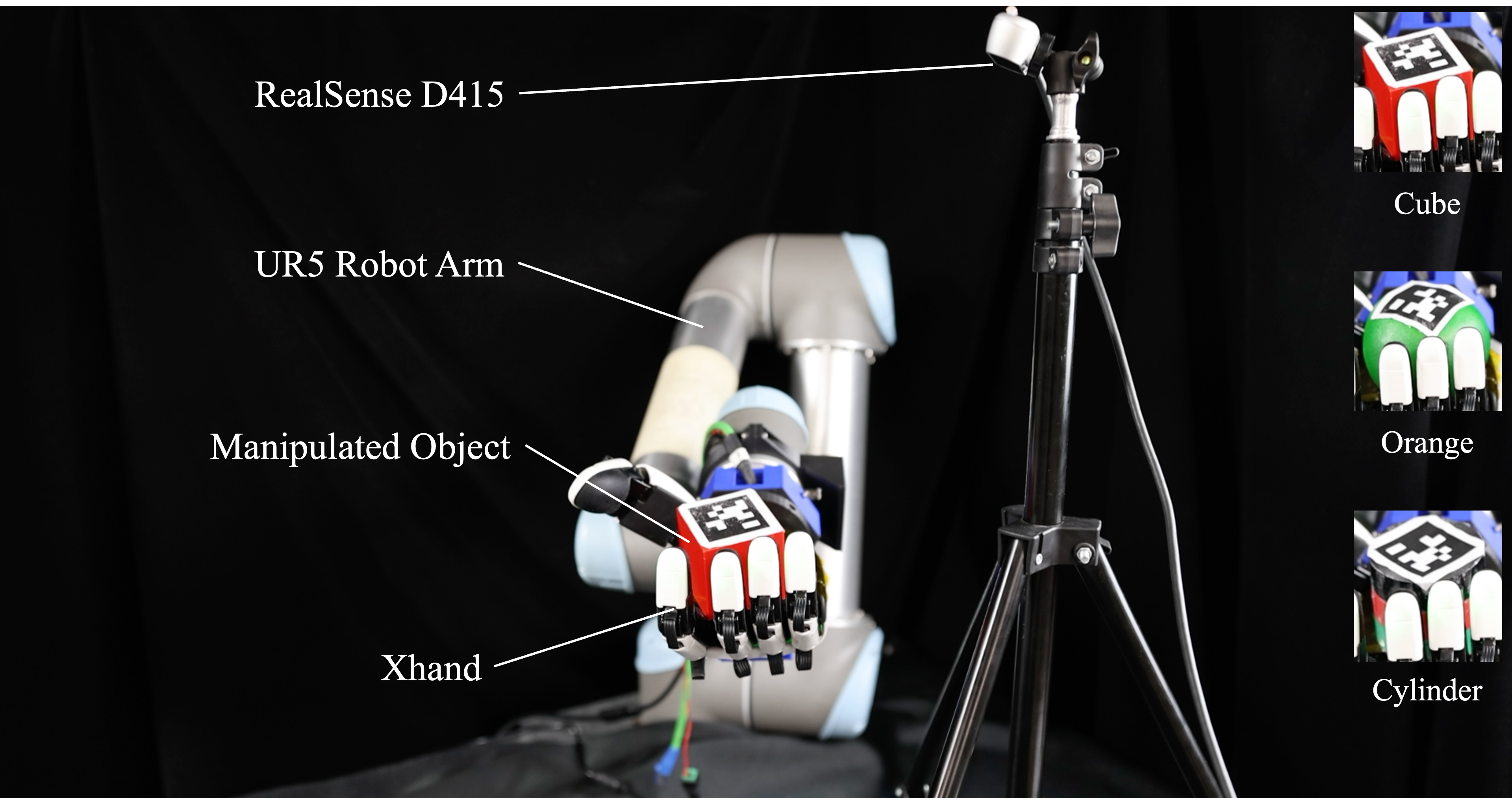}
\caption{\textbf{Experimental setup for real-world in-hand manipulation.}} 
\vskip -0.2in
\label{fig:inhand_setup}
\end{figure}

\begin{figure}[ht]
\centering
\includegraphics[width=0.92\textwidth]{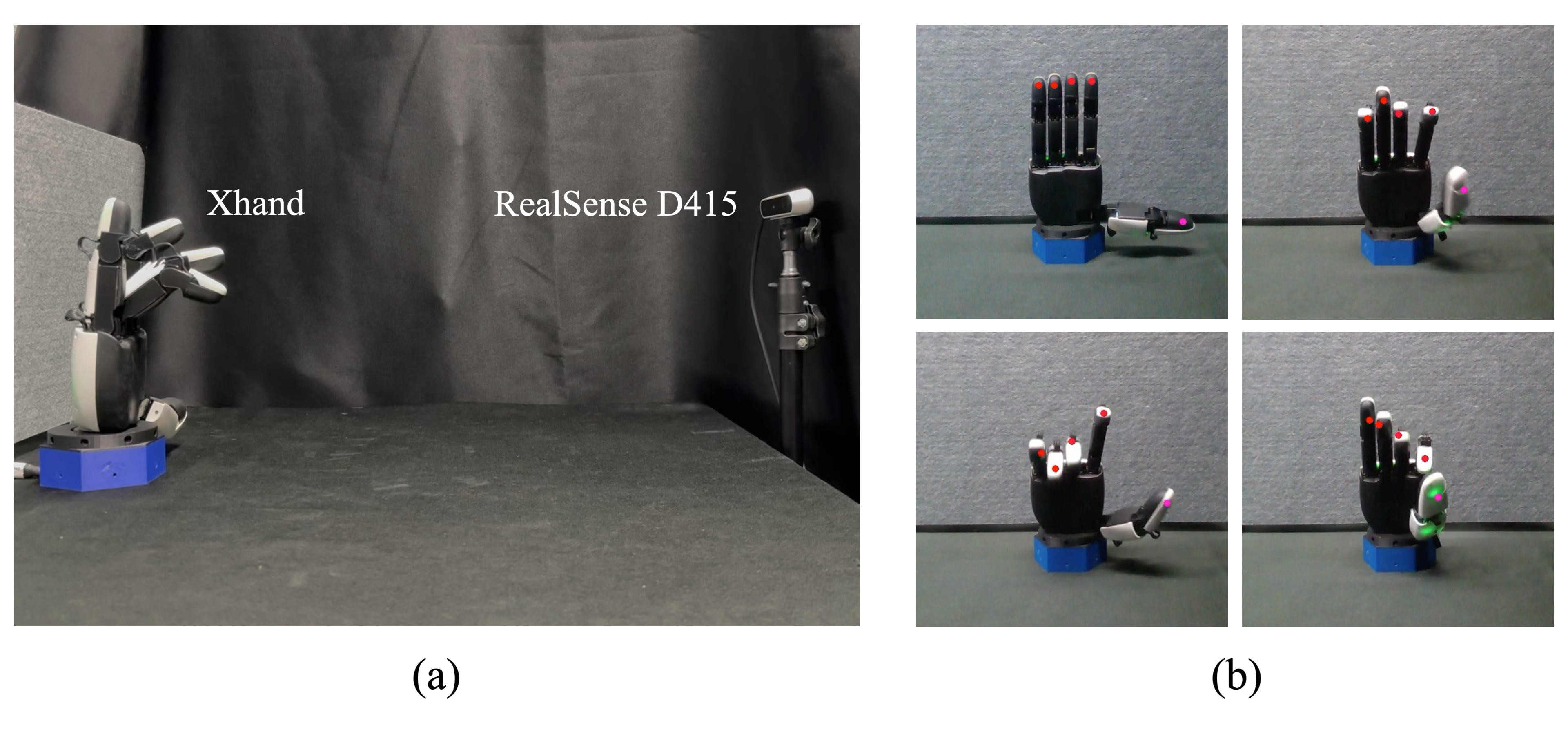}
\caption{\textbf{(a)} Experimental setup for real-world gesture generation. \textbf{(b)} 2D fingertip positions tracked over time using CoTracker3~\cite{karaev24cotracker3}.}
\vskip -0.2in
\label{fig:gesture_setup}
\end{figure}

We provide the hardware setup of in-hand manipulation and gesture generation in Figure~\ref{fig:inhand_setup} and Figure~\ref{fig:gesture_setup}(a) respectively.  

For in-hand manipulation, we represent the hand state using joint positions, while the object pose (position and orientation) is estimated using AprilTags detected by an Intel RealSense D415 camera. Object positions are computed in the image coordinate frame and normalized to the range $[0,1]$. We constrain orientation tracking to the rotation around the tag’s local $z$-axis. 

For gesture generation, we begin by recording video footage of the hand during exploration. We manually annotate three keypoints per finger across selected frames. These keypoints are then tracked throughout the video using CoTracker3~\cite{karaev24cotracker3}. To obtain a robust estimate of each finger’s position, we compute the average position of its tracked keypoints, weighting by their visibility confidence scores. The tracking results are shown in Figure~\ref{fig:gesture_setup}(b). We utilize the 2D positions in the image coordinate frame, as this information is sufficient to reliably infer diverse hand gestures given our setup. Notably, no camera-to-hand calibration is required for either in-hand manipulation or gesture generation, due to the design of our experimental setup.

\end{document}